\documentclass[11pt]{article}

\usepackage[preprint]{acl}

\usepackage{times}
\usepackage{latexsym}
\usepackage[T1]{fontenc}
\usepackage[utf8]{inputenc}
\usepackage{microtype}
\usepackage{inconsolata}
\usepackage{graphicx}

\usepackage{booktabs}
\usepackage{array}
\usepackage{supertabular}
\usepackage{placeins}
\usepackage{float}
\usepackage{pifont}
\usepackage{xspace}
\usepackage{amsmath,amssymb}


\newcommand{\NumReproduce}{63\xspace}
\newcommand{\NumDiscover}{43\xspace}
\newcommand{\NumProblems}{106\xspace}

\newcommand{\RepGptSub}{88.33}
\newcommand{\RepGptWhole}{68.25}

\newcommand{\RepKimiSub}{79.66}
\newcommand{\RepKimiWhole}{49.21}

\newcommand{\DisGlmSub}{59.34}

\newcommand{\DisQwenWhole}{18.60}

\newcommand{\DisToolGlmSub}{69.43}

\title{PHYRD: A Living Benchmark for Reproducing and Extending \\Theoretical Physics Research}
\author{\parbox{0.97\textwidth}{\centering\normalfont
\textbf{Qituan Zhang}\textsuperscript{1,2}\quad
\textbf{Guo Chen}\textsuperscript{1,3}\quad
\textbf{Linfeng Zhang}\textsuperscript{1,\ensuremath{\dagger}}\quad
\textbf{Wenlong Zhang}\textsuperscript{1,4}\\[0.7em]
{\small
\textsuperscript{1}EPIC Lab, Shanghai Jiao Tong University\\
\textsuperscript{2}Fudan University\quad
\textsuperscript{3}University of Electronic Science and Technology of China\\
\textsuperscript{4}University of Science and Technology of China\\[0.5em]
\textsuperscript{\ensuremath{\dagger}}Corresponding author:
\href{mailto:zhanglinfeng@sjtu.edu.cn}{\texttt{zhanglinfeng@sjtu.edu.cn}}
}}
}

\begin{document}
\raggedbottom
\maketitle

\begin{abstract}
Large language models and agents have shown significant power in advanced physical research. In this paper,
We introduce PHYRD, a living benchmark for theoretical-physics
reasoning with \NumReproduce Reproduce and \NumDiscover Discover
problems. Reproduce tests derivation of published results;
Discover tests consequences of modified theoretical conditions.
FormulaForest constructs the tasks with reference solutions and
subquestion-level rubrics. Across five models, whole-problem
accuracy ranges from \RepKimiWhole\% to \RepGptWhole\% on
Reproduce and from \DisQwenWhole\% to 74.42\% on closed-book
Discover. With tools, four models complete more Discover
problems, but conceptual errors and missing arguments persist. Besides, our experimental results demonstrate the value of FormulaForest as synthetic data. On FrontierScience, a benchmark developed by OpenAI, it raises the accuracy of a Qwen-35B-A3B model from 61\% to 72\%.

\end{abstract}

\section{Introduction}
\label{sec:introduction}
\begingroup
\newcommand{\phyrdYes}{\ding{51}}
\newcommand{\phyrdNo}{\ding{55}}
\newcommand{\phyrdNR}{\textemdash}

\begin{table*}[t]
\centering
\small
\renewcommand{\arraystretch}{1.12}
\begin{tabular*}{\textwidth}{@{\extracolsep{\fill}}lccccc@{}}
\toprule
Benchmark &
\shortstack{Physics\\focus} &
\shortstack{Tool\\use} &
\shortstack{Research\\tasks} &
\shortstack{Derivation under\\modified conditions} &
\shortstack{Continued\\expansion} \\
\midrule
MATH & \phyrdNo & \phyrdNo & \phyrdNo & \phyrdNo & \phyrdNo \\
ScienceAgentBench & \phyrdNo & \phyrdYes & \phyrdYes & \phyrdNR & \phyrdNR \\
PRL-Bench & \phyrdYes & \phyrdYes & \phyrdYes & \phyrdNR & \phyrdYes \\
HorizonMath & \phyrdNo & \phyrdNR & \phyrdYes & \phyrdNR & \phyrdYes \\
MathArena & \phyrdNo & \phyrdYes & \phyrdYes & \phyrdNR & \phyrdYes \\
LiveBench & \phyrdNo & \phyrdNo & \phyrdNo & \phyrdNo & \phyrdYes \\
\midrule
\textbf{PHYRD (ours)} & \phyrdYes & \phyrdYes & \phyrdYes & \phyrdYes & \phyrdYes \\
\bottomrule
\end{tabular*}
\caption{Coverage of representative benchmarks.
\phyrdYes: included; \phyrdNo: outside scope; \phyrdNR: not specified.
Modified-condition derivation examines consequences of changed premises;
continued expansion includes release plans for PRL-Bench and PHYRD.}
\label{tab:benchmark_comparison}
\end{table*}
\endgroup

Rapid advances in mathematical reasoning, code generation,
and scientific problem solving are expanding the potential
role of large language models (LLMs) in scientific research.
In theoretical physics, these capabilities could support
literature analysis, analytical derivation, computational
verification, and exploration of new theoretical settings.
Assessing whether models can reliably perform these
activities is an important question for AI-driven scientific research.

Existing benchmarks range from competition problem solving,
exemplified by MATH
\citep{hendrycks2021measuringmathematicalproblemsolving},
to scientific workflows and research-level tasks.
These settings assess different abilities and leave
distinct gaps in the evaluation of theoretical physics
research:

\noindent\textbf{Agentic Capabilities:}
Competition problems and static scientific question
answering offer limited evidence of how models use
external information and computational feedback to
advance a research task.
ScienceAgentBench extends evaluation to programming
and execution, focusing on generated programs and
their outputs
\citep{chen2025scienceagentbenchrigorousassessmentlanguage}.
In theoretical physics, successful execution or numerical
agreement alone does not establish the validity of an
analytical argument.
Evaluation must also examine the assumptions behind
formulas, the completeness of derivations, and whether
computational evidence supports a general conclusion.

\noindent\textbf{Frontier Exploration:}
Research-level difficulty does not by itself test
the ability to extend a theory.
PRL-Bench emphasizes open solution paths and
long-horizon research tasks, while HorizonMath
focuses on automatically verifiable open mathematical
problems
\citep{miao2026prlbenchcomprehensivebenchmarkevaluating,
wang2026measuringprogressreasoningmathematical}.
Neither organizes its principal task categories around
the distinction between deriving results under existing
conditions and deriving consequences after those
conditions change.
This distinction tests whether models can identify
invalidated arguments and derive new conclusions after
a model, assumption, or condition changes.

\noindent\textbf{Dynamic Updates:}
Fixed task collections can lose their ability to
distinguish models as capabilities improve, while
exposure to public questions and solutions during
training can compromise evaluation.
MathArena and LiveBench address these issues by
continually introducing new tasks
\citep{dekoninck2026benchmarksmatharenaevaluationplatform,
white2025livebenchchallengingcontaminationlimitedllm}.
Theoretical physics evaluation likewise needs to
follow research progress.
Each update also requires checking the theoretical
premises, reference derivations, and scoring criteria,
while retaining version records so that results
remain traceable.

Based on these observations, we introduce PHYRD, a living benchmark comprising
\NumReproduce Reproduce and \NumDiscover Discover problems.
The tracks assess derivation of published results and
analysis under modified theoretical conditions,
respectively. Discover specifies what changes relative
to the source study and which premises remain fixed.
Discover includes closed-book and tool-assisted
evaluation on the same problem set.
Each public subquestion is scored as correct or incorrect,
and we report both subquestion and whole-problem accuracy.
We will release timestamped additions drawn from newly
published research.
Table~\ref{tab:benchmark_comparison} compares the
task and design scope of representative benchmarks.

PHYRD is constructed using FormulaForest.
The framework organizes formulas, definitions,
and assumptions from source papers into mathematical
dependency relations with explicit applicability
conditions.
These relations guide the construction of problem
statements, reference solutions, and scoring criteria.
Reference derivations undergo source verification,
analytical review, and computational checks where
applicable. Section~\ref{sec:construction} and
Appendix~\ref{app:construction_pipeline} describe
the construction process.

Our experiments show that models can complete many local
derivations, but calculation errors and missing arguments
still prevent complete solutions. On Discover, four of
the five models solve more problems completely with tools;
DeepSeek and Qwen each gain 20.93 percentage points in
whole-problem accuracy. GLM improves in subquestion
accuracy but not in whole-problem accuracy.
Across paired Discover responses, calculation errors
decrease substantially in the tool-assisted setting,
whereas conceptual errors remain at a similar level.
The remaining failures concern theoretical structure
and missing arguments as well as algebraic operations.

\section{Related Work}
\label{sec:related_work}

\paragraph{Mathematical and physics benchmarks.}\enspace
MATH tests competition mathematics; SciBench and GPQA
assess scientific problems and expert knowledge
\citep{hendrycks2021measuringmathematicalproblemsolving,
wang2024scibenchevaluatingcollegelevelscientific,
rein2023gpqagraduatelevelgoogleproofqa}.
TheoremQA evaluates theorem application; FrontierMath
and HorizonMath test difficult or open mathematical problems
\citep{chen2023theoremqatheoremdrivenquestionanswering,
glazer2025frontiermathbenchmarkevaluatingadvanced,
wang2026measuringprogressreasoningmathematical}.
TPBench spans undergraduate and research physics;
PRL-Bench constructs research tasks from recent physics papers
\citep{chung2025theoreticalphysicsbenchmarktpbench,
miao2026prlbenchcomprehensivebenchmarkevaluating}.

\paragraph{Tool-assisted scientific research.}\enspace
ReAct combines reasoning with actions, while PAL
uses generated programs for calculation
\citep{yao2023reactsynergizingreasoningacting,
gao2023palprogramaidedlanguagemodels}.
SciCode evaluates scientific programming; ScienceAgentBench
and DiscoveryBench test data-driven workflows and hypothesis discovery
\citep{tian2024scicoderesearchcodingbenchmark,
chen2025scienceagentbenchrigorousassessmentlanguage,
majumder2024discoverybenchdatadrivendiscoverylarge}.
CORE-Bench and PaperBench assess computational reproduction
\citep{siegel2026corebenchfosteringcredibilitypublished,
starace2025paperbenchevaluatingaisability}. In contrast,
PHYRD focuses on analytical derivations under source
and modified theoretical conditions.

\paragraph{Updating and assessing reasoning.}\enspace
LiveBench and MathArena introduce new tasks over time
\citep{white2025livebenchchallengingcontaminationlimitedllm,
dekoninck2026benchmarksmatharenaevaluationplatform}.
ProcessBench tests error identification in stepwise mathematics
\citep{zheng2025processbenchidentifyingprocesserrors}.
Research derivations also require checking whether intermediate
results support the requested conclusions. Because model judgments
can be sensitive to response position and verbosity
\citep{zheng2023judgingllmasajudgemtbenchchatbot},
PHYRD uses explicit subquestion requirements and reports
complete solutions separately.

\paragraph{Training with generated derivations.}\enspace
The related study \citep{liu2026tokenalignmenteventcompletion}
uses SciDeriv from FormulaForest's Reproduce pipeline.
It introduces the ESCD objective for cross-tokenizer distillation. In contrast, this paper contributes physics task construction and evaluation.

\begin{figure*}[t]
    \centering
    \includegraphics[width=\textwidth]{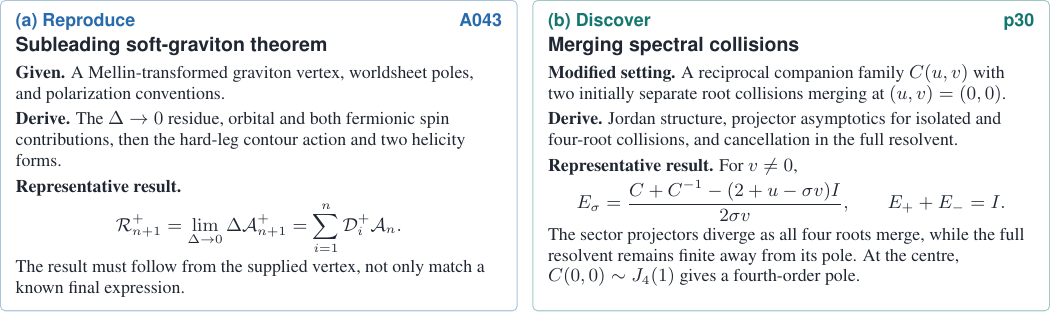}
    \caption{\textbf{Examples from the two tracks.}
    (a) Deriving a published soft graviton theorem.
    (b) Analyzing spectral collisions in a modified matrix family.
    Selected tasks and reference results are shown;
    complete statements and selected derivations appear
    in Appendix~\ref{app:reproduce_example}
    and~\ref{app:discover_example}.}
    \label{fig:task_examples}
\end{figure*}

\section{PHYRD Tasks and Dataset}
\label{sec:benchmark}
PHYRD comprises two tracks, Reproduce and Discover. We use the five subfields: astrophysics and
gravitation (Astro), condensed-matter physics (Cond-Mat), high-energy
physics (HEP), quantum physics and quantum information (Quantum),
and statistical physics (Stat). 
Each problem has a public statement, a reference solution,
per-subquestion scoring criteria, and source and verification
materials. Only the statement is given to the solver.

For example, celestial scattering amplitudes are assigned to HEP,
black-hole thermodynamics to Astro, and lattice many-body phases to
Cond-Mat. Quantum includes operator, tensor-network, and
quantum-information problems; Stat includes statistical mechanics,
stochastic dynamics, and thermodynamic constitutive relations.
For problems spanning several areas, the label records the primary
research context. Problem-level assignments are listed in
Section~\ref{app:catalogue}.

The \NumProblems problems contain 504 public subquestions:
279 in Reproduce and 225 in Discover. Their parts share
premises and may build on earlier results, so we assess
complete solutions and paired changes at the problem
level. Most Reproduce problems have four parts (39/63),
and most Discover problems have five (22/43); full size
distributions are in Appendix~\ref{app:problem_size}.

\subsection{Reproduce Tasks}
\label{sec:reproduce}

Reproduce asks for derivations of published formulas
and relationships under stated conditions. Correct
final expressions alone are insufficient: solutions must
handle conventions, normalization, and applicability.
Equivalent derivations are accepted.

Reproduce is closed book: models receive the public
statement but no source paper or external tools. In Figure~\ref{fig:task_examples}(a), the solver must
derive a soft-graviton theorem from a supplied vertex,
including its orbital and spin terms and the contour
action that produces the final expression.

\subsection{Discover Tasks}
\label{sec:discover}

Discover changes a theoretical object or condition
from a source paper. The public statement specifies
what changes, what remains fixed, and what to derive.
Models must check whether source relations still apply
and develop new arguments when they do not.
Discover extends the source setting without requiring
resolution of an open physics question.

Figure~\ref{fig:task_examples}(b) changes a matrix
family so that two root collisions merge. The solver
must derive the new Jordan and projector limits rather
than reuse an isolated-pair calculation.

We evaluate Discover both closed book and with specified
symbolic-computation and retrieval tools, on the same
\NumDiscover problems.

\subsection{Dataset and Updates}
\label{sec:dataset_updates}

We adopt the five-subfield taxonomy of
PRL-Bench~\citep{miao2026prlbenchcomprehensivebenchmarkevaluating}:
astrophysics, condensed matter physics, high-energy
physics, quantum information and foundations, and
statistical physics and complex systems.
Each problem receives one primary field by research
object and source context.

\begin{figure}[H]
    \centering
    \includegraphics[width=\columnwidth]{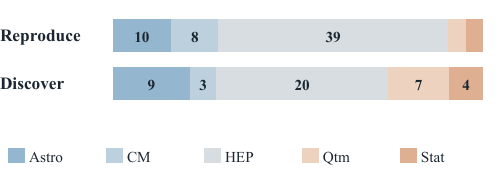}
    \caption{\textbf{Primary physics subfields in each track.} Counts
    are shown where segments permit; CM: condensed matter,
    Qtm: quantum.}
    \label{fig:field_mix}
\end{figure}

Figure~\ref{fig:field_mix} shows that HEP accounts for
39/63 Reproduce problems but 20/43 Discover problems.
Quantum and Stat together contribute six Reproduce
problems and eleven Discover problems. These counts describe the
dataset, not the breadth of the fields. Problem-level
assignments are in Appendix~\ref{app:catalogue}.

We plan timestamped additions from new papers;
each release will retain the prior task and score
versions needed to trace comparisons over time.

\section{FormulaForest Task Construction}
\label{sec:construction}
FormulaForest constructs tasks from relations in source
papers. Reproduce targets published results; Discover
modifies a local model or its conditions.

\subsection{Mathematical Relations}
\label{sec:relation_representation}

To retain definitions and validity conditions distributed
throughout a paper $s$, we represent its mathematical
content as a relation graph:
\begin{equation}
    \mathcal{G}_s = (V_s, E_s, \Pi_s),
    \label{eq:source_relation_graph}
\end{equation}
Here $V_s$ contains formulas and premises, $E_s$ records
derivation and specialization relations with their
conditions, and $\Pi_s$ records source locations.

Candidate relations are found through equation references
and context, then checked against the source and, where
needed, mathematically. An equation's position in a paper
does not establish a derivation, so explicit results and
inferred dependencies remain distinct.

\subsection{Task Construction}
\label{sec:dual_track_construction}

For Reproduce, we pair a published target with the
premises needed to derive it. The reference solution
fills gaps left by the source; scoring permits equivalent
arguments.

For Discover, a transformation is applied to a local
mathematical structure $H \subseteq \mathcal{G}_s$
and its conditions $C$:
\begin{equation}
    T : (H, C)
    \longmapsto
    (\widetilde{H}, \widetilde{C}).
    \label{eq:structural_transformation}
\end{equation}
Transformations may change operators, couplings,
symmetries, or boundary conditions. We record fixed
and changed premises, then determine which source
arguments survive and what new derivations are required.

Before release, derivations and conventions are verified;
Discover also checks that the modified model is well
defined. Blind solving probes whether the statement,
solution, and scoring criteria agree. The procedure is
in Appendix~\ref{app:construction_pipeline}.

The same pipeline produced the SciDeriv training corpus
used in a separate distillation study
\citep{liu2026tokenalignmenteventcompletion}; its results are
discussed in Section~\ref{sec:training_use}.

\section{Experiments}
\label{sec:experiments}
\subsection{Evaluation and Metrics}
\label{sec:experimental_setup}

Five models answer \NumReproduce Reproduce problems
without tools and the same \NumDiscover Discover problems
in closed-book and tool-assisted settings. Subquestion
accuracy (SAcc) averages each problem's correct-part
fraction; whole-problem accuracy (WAcc) requires every
part to be correct. Both use the full problem-set
denominator. Model settings, tool environments, answer
collection, and scoring rules are in
Appendix~\ref{app:evaluation}.

\subsection{Reproduce Results}
\label{sec:reproduce_results}

\begin{figure}[H]
    \centering
    \includegraphics[width=\columnwidth]{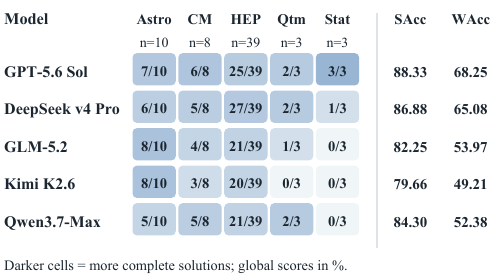}
    \caption{\textbf{Reproduce results on 63 problems}. Cells give complete
    solutions by primary subfield; the last two columns give global
    subquestion (SAcc) and whole-problem (WAcc) accuracy in percent.}
    \label{fig:reproduce_subfields}
\end{figure}

Figure~\ref{fig:reproduce_subfields} reports the Reproduce
scores. SAcc ranges from \RepKimiSub\% to \RepGptSub\%,
whereas WAcc ranges from \RepKimiWhole\% to \RepGptWhole\%.
GPT and DeepSeek solve 43 and 41 problems completely;
the other models solve 31--34.

Of 315 Reproduce responses, 182 are completely correct,
130 pass some but not all subquestions, and three pass
none. Among the 133 incomplete responses, 78 miss just
one subquestion. This concentration near completion
explains the gap between SAcc and WAcc: a missing
normalization, validity condition, or requested argument
still prevents a complete research result.

\begin{figure}[t]
\centering
\includegraphics[width=\columnwidth]{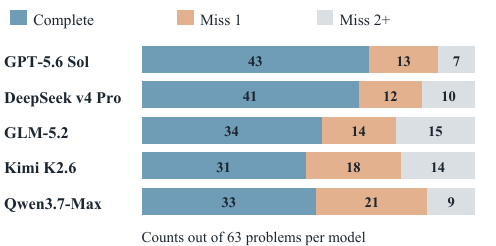}
\caption{Number of Reproduce responses by missed-subquestion
count. ``Miss 2+'' includes
responses with no correct subquestions.}
\label{fig:reproduce_distance}
\end{figure}

Figure~\ref{fig:reproduce_distance} separates complete answers
from one-part misses and larger failures. Qwen has 21
one-part misses, compared with 13 for GPT, but only 33
complete solutions against GPT's 43. The proximity of
an answer to completion therefore differs from the number
of finished derivations; neither is visible in SAcc alone.

Ten Reproduce problems are solved completely by all five
models, six by none, and ten by exactly one. The other
37 also have mixed outcomes. Aggregate rank therefore
does not capture which theoretical tasks each model solves.

The overall ordering varies across subfields
(Figure~\ref{fig:reproduce_subfields}). GLM and Kimi
each solve 8 of the 10 Astro problems, while DeepSeek
solves 27 of the 39 HEP problems, compared with 25
for GPT. HEP accounts for 61.9\% of Reproduce problems
and therefore contributes more to the overall score
than Astro. Quantum and Stat each contain only three
problems, so a single problem changes accuracy by
33.3 percentage points; these small subsets do not
support broad comparisons of domain-level ability.

\subsection{Discover: Closed Book and Tools}
\label{sec:discover_results}

\begin{figure}[t]
    \centering
    \includegraphics[width=\columnwidth]{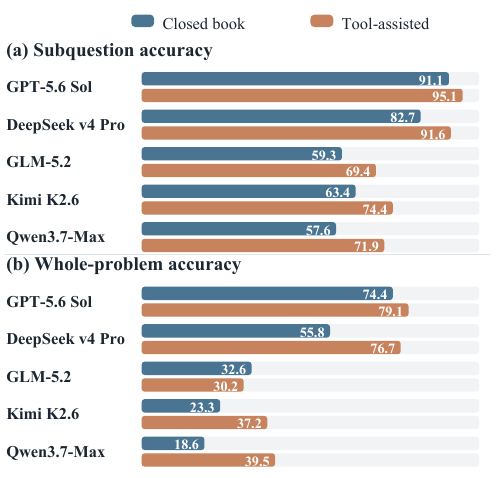}
    \caption{Discover accuracy on the same 43 problems in the two
    evaluated settings. Paired bars give closed-book (blue) and
    tool-assisted (orange) percentages.}
    \label{fig:discover_pairs}
\end{figure}

Figure~\ref{fig:discover_pairs} compares the two settings
on the same \NumDiscover Discover problems.
GPT, DeepSeek, Kimi, and Qwen gain 2, 9, 6, and 9
complete solutions, respectively; GLM loses one.

DeepSeek, Kimi, and Qwen gain 8.96, 11.02, and 14.30
SAcc points. Their paired problem-level 95\% bootstrap
intervals are [2.50, 16.54], [3.16, 18.96], and
[3.08, 25.55] points (10,000 resamples). The same
43 problems enter both settings, but the settings also
differ in interaction and computation, not only tool access.

GLM's SAcc rises from \DisGlmSub\% to \DisToolGlmSub\%,
but complete solutions fall from 14 to 13. On the 37
problems answered in both settings, SAcc barely changes
(68.96\% to 68.85\%), and the paired interval spans zero
([-10.52, 10.83] points). Improved answer coverage,
not better performance on those 37 tasks, drives the
full-set SAcc difference.

\begin{figure}[htb]
\centering
\includegraphics[width=\columnwidth]{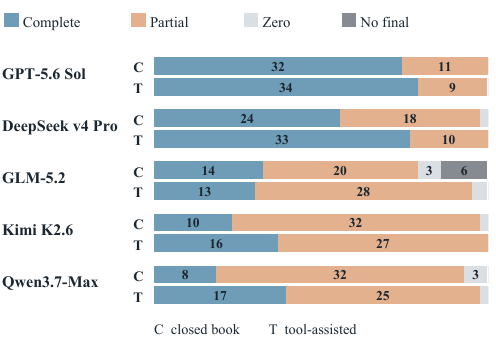}
\caption{Discover outcomes on the same 43 problems per model.
C: closed book; T: tool-assisted. Complete requires every
subquestion; partial passes at least one but not all.
The full 43-problem denominator is retained in both rows.}
\label{fig:discover_outcomes}
\end{figure}

The outcome breakdown in Figure~\ref{fig:discover_outcomes}
shows the coverage change directly: GLM has six problems
without a closed-book final answer and none with tools,
while its partial-answer count rises from 20 to 28 and
its complete-answer count falls from 14 to 13. DeepSeek's
complete count rises by nine as its partial count falls
from 18 to ten.

The SAcc--WAcc gap narrows for DeepSeek (26.86 to 14.89
points) but widens for GLM (26.78 to 39.20). More correct
parts coincide with complete solutions for the former,
but not the latter.

DeepSeek and Qwen each have nine more complete solutions
with tools, a net WAcc increase of 20.93 points.
Their SAcc gains are 8.96 and 14.30 points, respectively.
Tool-assisted WAcc reaches 76.7\% for DeepSeek and
39.5\% for Qwen, compared with GPT's 79.1\%.

\begin{figure}[H]
\centering
\includegraphics[width=\columnwidth]{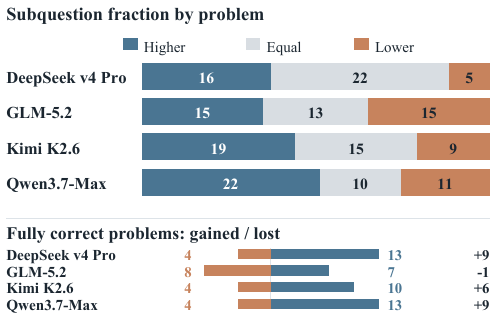}
\caption{Problem-level Discover changes for four API models
on 43 paired tasks each. Upper: subquestion fraction higher,
equal, or lower with tools. Lower: complete solutions gained
(blue), lost (orange), and net change (right).}
\label{fig:paired_changes}
\end{figure}

Figure~\ref{fig:paired_changes} reveals gains and losses
behind the averages. DeepSeek and Qwen each gain 13
complete solutions but lose four; GLM gains seven and
loses eight. More problems improve than decline for
DeepSeek, Kimi, and Qwen, while GLM has 15 in each
direction. Qwen's larger SAcc gain still leaves its
tool-assisted score below DeepSeek.

\begin{figure}[htb]
\centering
\includegraphics[width=\columnwidth]{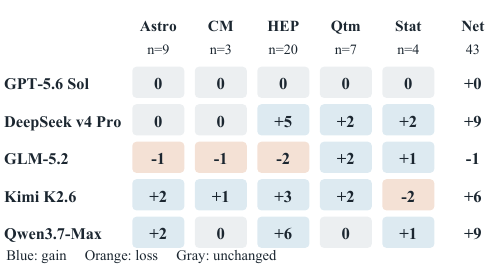}
\caption{Change in complete Discover solutions with tools,
by primary subfield (tool-assisted minus closed-book).
Column headings give the number of problems; Net sums
the five fields. Color distinguishes gains, losses, and ties.}
\label{fig:discover_field_changes}
\end{figure}

Figure~\ref{fig:discover_field_changes} locates the
changes: DeepSeek and Qwen gain five and six complete
HEP solutions, while GLM loses two in HEP and gains
two in Quantum. The small non-HEP subsets permit
coverage counts, not stable domain rankings. Full scores
are in Appendix~\ref{app:subfield_results}.

\begin{figure}[htb]
\centering
\includegraphics[width=\columnwidth]{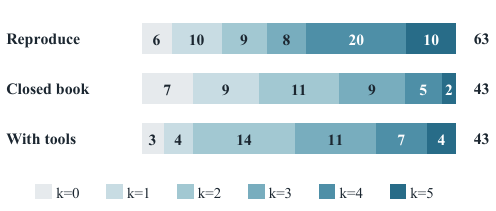}
\caption{Problems solved completely by exactly $k$ of five models.
Segment widths give within-setting shares; segment labels give problem
counts. The rightmost numbers are the setting totals.}
\label{fig:joint_solutions}
\end{figure}

In Figure~\ref{fig:joint_solutions}, Discover problems
solved by no model fall from seven to three with tools,
and those solved by all five rise from two to four.
Yet the largest tool-assisted group consists of 14
problems solved by exactly two models: broader coverage
does not mean consensus on difficult tasks.

\subsection{Error Analysis}
\label{sec:error_analysis}

\begin{figure*}[!t]
\centering
\includegraphics[width=0.97\textwidth]{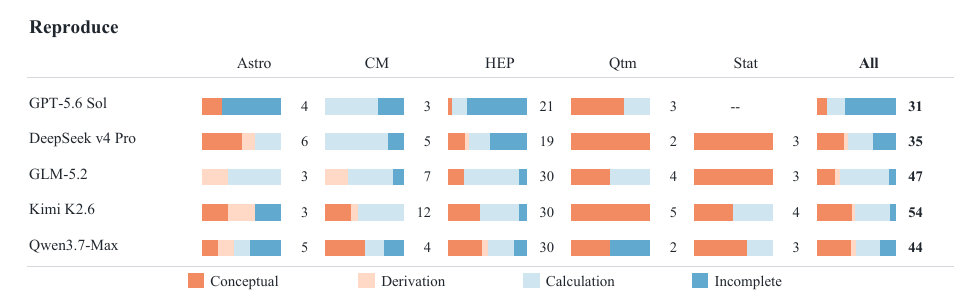}
\caption{Reproduce error composition by model and subfield.
Colored segments give the four category shares among errors;
numbers at right are counts. CM: condensed matter;
Qtm: quantum; \emph{All} pools subfields.}
\label{fig:error_reproduce_fields}
\end{figure*}

\begin{figure*}[!t]
\centering
\includegraphics[width=0.97\textwidth]{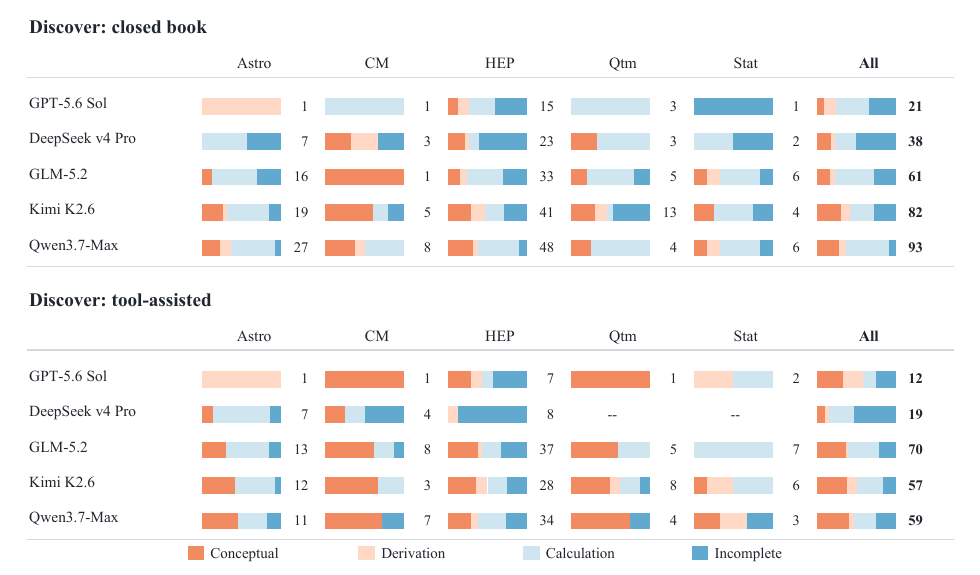}
\caption{Discover error composition: closed-book (top) and
tool-assisted (bottom). Segment widths and counts follow
Figure~\ref{fig:error_reproduce_fields}.}
\label{fig:error_discover_fields}\vspace{-0.25cm}
\end{figure*}

Figures~\ref{fig:error_reproduce_fields}
and~\ref{fig:error_discover_fields} show the composition
of incorrect subquestions by model and subfield. Each
receives one primary label: conceptual, derivation,
calculation, or incomplete or unsupported. The bars
show shares \emph{among errors}, not error rates; counts
appear to their right. Appendix~\ref{app:error_coding}
gives the exact counts and representative cases.

On Reproduce, incomplete derivations or conclusions
account for 20 of GPT's 31 incorrect subquestions,
whereas calculation errors account for 29 of GLM's 47.
The former include stating a result without the
independent derivation requested by the problem;
the latter include sign and coefficient errors.
These different error profiles help explain why similar
overall scores need not indicate the same limitations.
A computational check can address a wrong coefficient,
whereas an omitted argument still has to be supplied
and justified under the problem's assumptions.
The distinction remains within HEP, the largest
Reproduce subfield: 16 of GPT's 21 errors there are
incomplete or unsupported, while 21 of GLM's 30 are
calculation errors. Their different overall profiles
are therefore not simply a consequence of answering
different proportions of physics subfields. Small
subfields should be read by the counts beside each bar:
a single error can occupy its entire bar.

For Discover, we compare error counts on
the same answered problems in both settings: 37 for
GLM and 43 for each other model. Across these 209
answer pairs, calculation errors decrease from 123
without tools to 67 with tools, while conceptual errors
change from 70 to 75. Incomplete or unsupported
subquestions decrease from 74 to 50, and derivation
errors from 28 to 16. These are counts of incorrect
subquestions, not just percentages of a shrinking error
pool. Of the net 87 fewer incorrect parts in this paired
set, 56 come from the calculation category. Conceptual
errors rise by five in absolute count, so their larger
share among remaining errors is not only a denominator
effect.

Figure~\ref{fig:error_discover_fields} shows where the
remaining Discover errors lie. GLM has 26 conceptual and
27 calculation errors with tools. Qwen's calculation
errors fall from 50 to 16, but conceptual errors move
only from 26 to 24. DeepSeek has 19 tool-assisted errors,
ten of them incomplete or unsupported. Thus fewer
calculation mistakes need not resolve errors in premises,
scope, or required conclusions.

This pattern is not uniform across models. In HEP,
which contains 20 Discover problems, the raw error
counts fall from 23 to eight for DeepSeek and from 48
to 34 for Qwen, but rise from 33 to 37 for GLM.
As with the overall GLM result, the latter comparison
also reflects different answer coverage across settings.
The field-level display therefore supports the aggregate
contrast without implying uniform gains.

DeepSeek and Qwen each finish with nine more fully
correct Discover problems, although their incorrect-part
counts fall by 19 (38 to 19) and 34 (93 to 59), respectively.
GPT has nine fewer incorrect parts (21 to 12) but only
two more complete problems. GLM's error count rises
from 61 to 70 as final-answer coverage improves, while
its number of complete solutions falls from 14 to 13. 
Each incorrect subquestion has one primary error label,
although its grading record may identify additional
defects. The paired counts describe the evaluated
responses; the different tool environments and reasoning
budgets do not isolate a tool-access effect.

\subsection{Training with Generated Tasks}
\label{sec:training_use}

FormulaForest's Reproduce pipeline also produced the
SciDeriv training corpus used in a separate cross-tokenizer
distillation study~\citep{liu2026tokenalignmenteventcompletion}.
Its retained 17,577 prompts comprise 10,000 conclusion-assessment
and 7,577 conclusion-reconstruction tasks. The former ask a
student whether a candidate claim follows from, is merely
compatible with, or conflicts with the given relations. The
latter withhold the endpoint and require the strongest conclusion
supported by the derivation. Both exercise the relation between
theoretical conditions and claims, rather than recalling
isolated formulas.
SciDeriv contains prompts but no stored reference
completions or scalar reward labels, which are acceptable for on-policy distillation.

\begin{figure}[htb]
\centering
\includegraphics[width=\columnwidth]{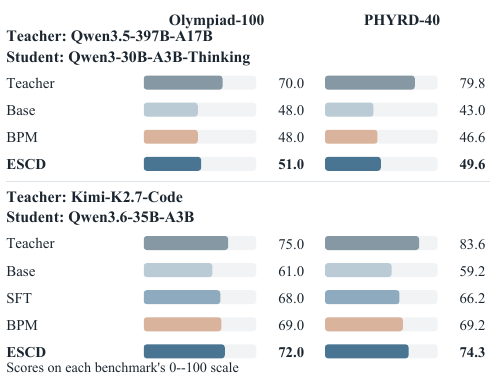}
\caption{External distillation results on Olympiad-100 (a subset from FrontierScience benchmark, made by OpenAI)
(accuracy, \%) and PHYRD-40 (graded mean, 0--100).
Teacher rows are references; BPM and ESCD are two on-policy distillation methods that use SciDeriv
prompts, with the 35B student initialized by SFT~\cite{liu2026tokenalignmenteventcompletion}.}
\label{fig:training_use}
\end{figure}

The study evaluates two teacher--student pairs on PHYRD-40,
a separate 40-problem physics set scored from 0 to 100,
and on FrontierScience Olympiad-100. Figure~\ref{fig:training_use}
shows the teacher references, base students, a token-alignment
baseline (BPM), and event-set completion distillation (ESCD).
The 35B student also has a supervised fine-tuning (SFT)
checkpoint from which both BPM and ESCD start. The 30B
student has no SFT stage. PHYRD-40's graded mean scores
are not directly comparable to our binary PHYRD scores.

With the 397B Qwen teacher and 30B student, ESCD moves
PHYRD-40 from 43.0 to 49.6 and Olympiad-100 from 48.0
to 51.0. It exceeds BPM by 3.0 points on each benchmark.
With the 1T Kimi teacher and 35B student, SFT first raises
the base scores from 59.2 to 66.2 and from 61.0 to 68.0.
From that same SFT checkpoint, ESCD reaches 74.3 on
PHYRD-40 and 72.0 on Olympiad-100, exceeding BPM by
5.1 and 3.0 points, respectively. The gains thus differ
across the two teacher--student pairs and benchmarks;
the larger pair also uses a different initialization.

Both BPM and ESCD use SciDeriv prompts. Their comparison
concerns the distillation objective, while the base-to-ESCD
change also includes training and, for the 35B student,
SFT. The reported teacher scores are 79.8 and 83.6 on
PHYRD-40, leaving gaps of 30.2 and 9.3 points after ESCD.
The same ordering appears on Olympiad-100: the ESCD
students score 51.0 and 72.0 against teacher references
of 70.0 and 75.0. These results show a downstream use of
the generated derivation tasks beyond benchmark evaluation.

\FloatBarrier

\section{Conclusion}
\label{sec:conclusion}
Across both tracks in PHYRD, models often complete individual
derivations without satisfying every requirement of a
research problem. On Discover, four of the five models
solve more problems completely in the tool-assisted
setting, yet failures remain when a calculation is applied
outside theoretical conditions. 

\section*{Limitations}
Uneven subfield coverage limits domain comparisons.
Models may have encountered related published results
during training. 

\section*{Ethics Statement}
All evidence comes from publicly available sources. We used AI assistants for language editing and coding support; the authors reviewed all content and take full responsibility.

\clearpage
\flushbottom
\bibliography{references}

\appendix
\section{Dataset and Task Examples}
\label{app:dataset}

This appendix describes the subfield assignments and problem sizes,
presents one Reproduce example and one Discover example, and lists
all 106 problems with their sources. The examples include complete
statements and selected derivations.

\subsection{Primary physics subfields}
\label{app:subfields}

We use the five subfields adopted in the main text: astrophysics and
gravitation (Astro), condensed-matter physics (Cond-Mat), high-energy
physics (HEP), quantum physics and quantum information (Quantum),
and statistical physics (Stat). Each problem is assigned one primary
field according to its research object and source context, rather
than its mathematical methods alone.

For example, celestial scattering amplitudes are assigned to HEP,
black-hole thermodynamics to Astro, and lattice many-body phases to
Cond-Mat. Quantum includes operator, tensor-network, and
quantum-information problems; Stat includes statistical mechanics,
stochastic dynamics, and thermodynamic constitutive relations.
For problems spanning several areas, the label records the primary
research context. Problem-level assignments are listed in
Section~\ref{app:catalogue}; the subfield distribution is shown
in the main text.

\subsection{Problem size and shared context}
\label{app:problem_size}

Reproduce contains 63 problems with 279 public subquestions, and
Discover contains 43 problems with 225 public subquestions.
Table~\ref{tab:app_subquestion_distribution} gives the distribution
of problem sizes. Most problems have four or five subquestions.

Subquestions within a problem share definitions and assumptions.
They may ask for successive steps in a derivation or examine
different consequences of the same theoretical setup. For example,
the Discover problem in Section~\ref{app:discover_example} uses one
matrix family to study eigenvalues, Jordan structure, spectral
projectors, and resolvents. This grouping is preserved when
computing problem-level scores and whole-problem accuracy.

\begin{table*}[t]
\centering
\small
\begin{tabular*}{\textwidth}{@{\extracolsep{\fill}}lrrrrrrrr@{}}
\toprule
Track & 3 & 4 & 5 & 6 & 7 & 8 & 9 & Total \\
\midrule
Reproduce & 1 & 39 & 20 & 2 & 0 & 1 & 0 & 63 \\
Discover & 3 & 6 & 22 & 7 & 2 & 1 & 2 & 43 \\
\bottomrule
\end{tabular*}
\caption{Number of problems with each subquestion count.
Column headings 3--9 denote the number of public subquestions
in a problem.}
\label{tab:app_subquestion_distribution}
\end{table*}

\subsection{Soft-graviton example (A043)}
\label{app:reproduce_example}

Problem A043 asks for the subleading conformal soft graviton
theorem starting from a supplied graviton vertex. Its source
is R07 in Table~\ref{tab:app_source_catalog}.

\subsubsection{Problem statement}

Consider the ambitwistor-string representation of gravitational
scattering in a celestial conformal-primary basis.
Write a hard null momentum as
$p_i^\mu=\omega_iq_i^\mu$. A soft graviton has null direction
$q^\mu$, energy $\omega$, and factorized polarization
\begin{align}
\varepsilon_{\mu\nu}&=\varepsilon_\mu\varepsilon_\nu,
&q^2&=q\cdot\varepsilon=0,\nonumber\\
\varepsilon_\mu&\sim\varepsilon_\mu+\alpha q_\mu.
\end{align}
The worldsheet fields are $X^\mu$, its conjugate momentum $P_\mu$,
and fermions $\Psi^\mu,\widetilde\Psi^\mu$. Antisymmetrization is
defined without a factor of one half:
\begin{equation}
A^{[\mu}B^{\nu]}=A^\mu B^\nu-A^\nu B^\mu.
\end{equation}
The pre-soft integrated insertion supplied to the solver is
\begin{align}
\mathcal V^{\rm grav}_{\Delta,\varepsilon}
={}&\int_0^\infty d\omega\,\omega^{\Delta-1}
\int_\Sigma\bar\delta(\omega q\cdot P)
\nonumber\\
&\quad\times\varepsilon_{\mu\nu}
\bigl(P^\mu+i\omega q_\rho\Psi^\rho\Psi^\mu\bigr)
\nonumber\\
&\quad\times
\bigl(P^\nu+i\omega q_\sigma\widetilde\Psi^\sigma
                 \widetilde\Psi^\nu\bigr)e^{i\omega q\cdot X}.
\end{align}
Near a hard insertion, the local behavior is
\begin{align}
P^\mu(\sigma)V_i(\sigma_i)&\sim
 \frac{p_i^\mu}{\sigma-\sigma_i}V_i(\sigma_i),\\
q\cdot P(\sigma)V_i(\sigma_i)&\sim
 \frac{q\cdot p_i}{\sigma-\sigma_i}V_i(\sigma_i).
\end{align}
For the four-dimensional expression one may use
\begin{align}
q^\mu(z,\bar z)&=
 (1+z\bar z,\ z+\bar z,\nonumber\\
&\qquad -i(z-\bar z),\ 1-z\bar z),\nonumber\\
q_i^\mu&=q^\mu(z_i,\bar z_i).
\end{align}
with $h_i=(\Delta_i+J_i)/2$ and
$\bar h_i=(\Delta_i-J_i)/2$.

Starting from this vertex, derive the subleading conformal soft
graviton theorem at $\Delta\to0$:
\begin{enumerate}
\item Isolate the $\Delta\to0$ residue of the graviton vertex.
\item Rewrite it in terms of orbital and spin angular-momentum currents.
\item Evaluate its insertion in an $n$-point hard celestial correlator
by deforming the contour around the hard punctures.
\item Give the covariant theorem and both the positive- and
negative-helicity forms in four-dimensional celestial coordinates.
\end{enumerate}

\subsubsection{Selected derivation}

\paragraph{Extracting the soft residue.}
Using $\bar\delta(\omega q\cdot P)=\omega^{-1}\bar\delta(q\cdot P)$,
expand the numerator as $B_0+i\omega B_1+O(\omega^2)$, where
\begin{align}
B_0={}&(\varepsilon\cdot P)^2,\\
B_1={}&(q\cdot X)(\varepsilon\cdot P)^2
 +(\varepsilon\cdot P)(q\cdot\Psi)(\varepsilon\cdot\Psi)
\nonumber\\
& +(\varepsilon\cdot P)(q\cdot\widetilde\Psi)
   (\varepsilon\cdot\widetilde\Psi).
\end{align}
The small-energy Mellin integrals, continued meromorphically from
their respective convergence domains, distinguish the two poles:
\begin{align}
\int_0^\Lambda\omega^{\Delta-2}d\omega
&=\frac{\Lambda^{\Delta-1}}{\Delta-1},\nonumber\\
\int_0^\Lambda\omega^{\Delta-1}d\omega
&=\frac{\Lambda^\Delta}{\Delta}.
\end{align}
Thus the constant term produces the leading pole at $\Delta=1$,
whereas the term linear in energy gives the requested residue:
\begin{equation}
\operatorname*{Res}_{\Delta=0}
\mathcal V^{\rm grav}_{\Delta,\varepsilon}
=i\int_\Sigma\bar\delta(q\cdot P)\,B_1.
\end{equation}
Both fermionic factors contribute at this order.

\paragraph{Current and hard-leg action.}
The orbital identity
\begin{equation}
\varepsilon_\mu q_\nu P^{[\mu}X^{\nu]}
=(\varepsilon\cdot P)(q\cdot X)
 -(q\cdot P)(\varepsilon\cdot X)
\end{equation}
uses the specified antisymmetrization convention. The second term
vanishes on the support of $\bar\delta(q\cdot P)$.
The two fermion bilinears supply spin contributions; their ordering
must be tracked when rewriting them as currents.

The localized insertion is then expressed as a contour integral
and deformed to the hard punctures. The orbital and spin terms act
together as the total angular-momentum generator on each hard state.
Substituting the four-dimensional null directions and polarization
vectors gives differential operators in $z_i,\bar z_i$ for the
two helicities.

The grading follows the four requests in the statement. In particular,
a correct soft factor alone does not establish the contour action or
both helicity forms; consistent alternative polarization and overall
phase conventions are accepted.

\subsection{Discover example: root collisions (p30)}
\label{app:discover_example}

Problem p30 studies how spectral projectors change when an isolated
two-root collision becomes a simultaneous four-root collision.
It asks for the projectors, their singular limits, and the associated
Jordan structure. Its source context is listed in
Table~\ref{tab:app_discover_catalog}.

\subsubsection{Problem statement}

Here reciprocal means a palindromic characteristic polynomial with
determinant one; no symplectic-form constraint is imposed. For a real
$4\times4$ matrix $T$, define
\begin{align}
c_1&=\operatorname{tr}T,\nonumber\\
c_2&=\tfrac12\bigl((\operatorname{tr}T)^2
  -\operatorname{tr}(T^2)\bigr),\nonumber\\
D&=c_1^2-4(c_2-2).
\end{align}
The problem has five public parts.

\paragraph{1. Spectral data and Jordan structure.}
Reduce $\chi_T(z)$ using $\eta=z+z^{-1}$. Compare the Jordan structures,
nullity ladders, and growth of powers for
\begin{gather}
\operatorname{diag}(g,g,g^{-1},g^{-1}),\qquad
J_2(g)\oplus J_2(g^{-1}),\\
I_4,\qquad J_2(1)\oplus J_2(1),\qquad J_4(1),
\end{gather}
where $g\in\mathbb R\setminus\{0,1,-1\}$. Prove that the characteristic
polynomial and nullity ladders at every complex root determine the
real similarity class. Determine the smallest uniform number of
ladder entries needed in dimension four, and prove both necessity
and sufficiency. Explain the reciprocal and complex-conjugacy
restrictions, including which partitions are possible at a sole
self-reciprocal root.

\paragraph{2. Root collisions and analytic arcs.}
Introduce the companion family
\begin{equation}
C(u,v)=
\begin{pmatrix}
0&0&0&-1\\
1&0&0&4+2u\\
0&1&0&-(6+4u+u^2-v^2)\\
0&0&1&4+2u
\end{pmatrix}.
\end{equation}
Factor its characteristic polynomial into two quadratics labelled
$\sigma=\pm1$, with $\delta_\sigma=u+\sigma v$. Describe the local
discriminant near $(0,0)$, excluding remote band edges. For analytic
arcs
\begin{equation}
\delta_\sigma(t)=a_\sigma t^{r_\sigma}
                 +O(t^{r_\sigma+1}),\qquad a_\sigma\ne0,
\end{equation}
derive the leading displacements of the roots from 1 and a valid
remainder order. Give arcs with root-valuation pairs $(2,6)$ and
$(4,4)$. In complexified parameter space, determine the permutations
induced by band-edge meridians. State what finite root jets cannot
establish.

\paragraph{3. An isolated double-root limit.}
Compute the complete adjugate
$A(x)=\operatorname{adj}(xI-C)$ and the simple-root projectors.
On $v=0$, at fixed $u\notin\{0,-4\}$, derive the Laurent matrices
$N_z,E_z$ at either double root $z$, in terms of
$A(z),A'(z)$, and $z-z^{-1}$. Prove their ranks, nilpotency,
projector identities, and the finite sum rule when two isolated
simple roots approach this double root.

\paragraph{4. A simultaneous four-root limit.}
Now let $(u,v)\to(0,0)$ through configurations with four simple roots
and $v\ne0$. Let $E_\sigma$ be the sum of the two projectors associated
with the quadratic labelled $\sigma$. Express $E_\sigma$ as a
polynomial or rational function of $C,u,v$, and derive each individual
projector from it. Determine whether $E_\sigma$ remains finite.
Derive its leading matrix asymptotic and that of the individual
projectors along the arcs above. Explain the algebraic cancellations
in the full resolvent for fixed $x$ away from 1. An isolated-pair
limit must not be assumed to remain valid when all four roots merge.

\paragraph{5. Resolvent orders and inverse identification.}
Compare the resolvent pole orders and Laurent-coefficient ranks at
$I_4$, $J_2(1)\oplus J_2(1)$, and $C(0,0)$. Relate them to nullity
ladders and finite-cell growth. Give an inverse protocol based on
unordered spectral data and matrix ranks, and identify the smallest
rank datum that separates these three centers. No versal-deformation
or material-specific claim is requested.

\subsubsection{Selected derivation}

\paragraph{Eigenvalues do not determine the Jordan blocks.}
The palindromic polynomial reduces through $\eta=z+z^{-1}$:
\begin{align}
\chi_T(z)&=z^4-c_1z^3+c_2z^2-c_1z+1,\nonumber\\
z^{-2}\chi_T(z)&=\eta^2-c_1\eta+c_2-2.
\end{align}
The diagonal matrix $\operatorname{diag}(g,g,g^{-1},g^{-1})$
and $J_2(g)\oplus J_2(g^{-1})$ have the same characteristic
polynomial for real $g\notin\{0,\pm1\}$. At either eigenvalue their
nullity ladders are $(2,2,2,2)$ and $(1,2,2,2)$.
For $d_j=\dim\ker(T-zI)^j$, the difference $d_j-d_{j-1}$ counts
Jordan blocks of size at least $j$. At multiplicity four, the
partitions $1^4,2+1+1,2+2,3+1,4$ give first-two nullities
\begin{equation}
(4,4),\ (3,4),\ (2,4),\ (2,3),\ (1,2),
\end{equation}
respectively. Two entries suffice when the characteristic polynomial
is known; one does not distinguish $2+2$ from $3+1$. Conjugate
eigenvalues of a real matrix have matching Jordan partitions.
Reciprocal eigenvalues need equal algebraic multiplicities, but
spectral reciprocity alone does not impose matching partitions.

\paragraph{The modified companion family.}
Its characteristic polynomial is $q_+(z)q_-(z)$, where
\begin{align}
q_\sigma(z)&=z^2-(2+\delta_\sigma)z+1,\nonumber\\
\delta_\sigma&=u+\sigma v,\quad\sigma\in\{+1,-1\}.
\end{align}
The exact roots are
\begin{align}
z_{\sigma,\epsilon}
&=1+\frac{\delta_\sigma}{2}
 +\frac{\epsilon}{2}\sqrt{\delta_\sigma(4+\delta_\sigma)},\nonumber\\
\epsilon&\in\{+1,-1\}.
\end{align}
Near the origin, the collision components are
$\delta_+=0$, $\delta_-=0$, and $v=0$; the remote quadratic edges
$\delta_\sigma=-4$ are outside this local neighborhood.
For $a_\sigma\ne0$ and $r_\sigma\ge1$, on a chosen branch along an analytic arc
$\delta_\sigma(t)=a_\sigma t^{r_\sigma}+O(t^{r_\sigma+1})$,
\begin{equation}
z_{\sigma,\epsilon}-1
=\epsilon\sqrt{a_\sigma}\,t^{r_\sigma/2}
 +O\!\left(t^{\min(r_\sigma,r_\sigma/2+1)}\right).
\end{equation}
The remainder cannot in general be replaced by $O(t^{r_\sigma})$:
the next term inside $\delta_\sigma$ also enters its square root.
The arcs $(\delta_+,\delta_-)=(t^4,t^{12})$ and $(t^8,2t^8)$
give root-displacement orders $(2,6)$ and $(4,4)$.
Meridians around the two local quadratic edges interchange the
two roots within their respective sectors. A finite root expansion
cannot prove that $v$ vanishes identically along an analytic arc.

\paragraph{Adjugate and an isolated double-root limit.}
Put $a=4+2u$, $b=6+4u+u^2-v^2$. Write the adjugate as
$A(x)=(a_1\ a_2\ a_3\ a_4)$, where
\begin{align}
a_1&=\begin{pmatrix}x^3-ax^2+bx-a\\x^2-ax+b\\x-a\\1\end{pmatrix},
\nonumber\\
a_2&=\begin{pmatrix}-1\\x(x^2-ax+b)\\x(x-a)\\x\end{pmatrix},
\nonumber\\
a_3&=\begin{pmatrix}-x\\ax-1\\x^2(x-a)\\x^2\end{pmatrix},
\nonumber\\
a_4&=\begin{pmatrix}-x^2\\x(ax-1)\\ax-bx^2-1\\x^3\end{pmatrix}.
\end{align}
Multiplication verifies $(xI-C)A(x)=\chi_C(x)I$.
At a simple root $z$, $P_z=A(z)/\chi_C'(z)$.
On $v=0$, fix $u\notin\{0,-4\}$ and let $z,w=z^{-1}$
be the two distinct roots of the common quadratic. Expanding
$A(x)/((x-z)^2(x-w)^2)$ around $z$ gives
\begin{align}
N_z&=\frac{A(z)}{(z-w)^2},\nonumber\\
E_z&=\frac{A'(z)}{(z-w)^2}-\frac{2A(z)}{(z-w)^3}.
\end{align}
The companion matrix is cyclic and has one size-two Jordan block
at each root. Thus $N_z$ has rank one and $E_z$ rank two, with
$N_z^2=0$, $E_z^2=E_z$, and $E_zN_z=N_zE_z=N_z$.
When two isolated simple roots approach $z$, their projector sum
tends to $E_z$, even though the individual projectors diverge.

\paragraph{Why the four-root limit is different.}
For four simple roots and $v\ne0$, write
$H=C+C^{-1}$ and $h_\sigma=2+u+\sigma v$. The two sector projectors
and the individual projectors are
\begin{align}
E_\sigma&=\frac{H-h_{-\sigma}I}{2\sigma v},\\
P_{\sigma,\epsilon}
&=\frac{(C-z_{\sigma,-\epsilon}I)E_\sigma}
 {z_{\sigma,\epsilon}-z_{\sigma,-\epsilon}}.
\end{align}
The polynomial identity
$(H-h_+I)(H-h_-I)=0$ establishes idempotence and
$E_++E_-=I$. Let $K=C(0,0)-I$. Then $K^4=0$, $K^3\ne0$, and
\begin{align}
H(0,0)-2I&=K^2-K^3=:B_0,\nonumber\\
\operatorname{rank}B_0&=2.
\end{align}
Consequently, as both parameters tend to zero,
\begin{align}
E_\sigma&\sim\frac{B_0}{2\sigma v},\nonumber\\
P_{\sigma,\epsilon}&\sim
\frac{\epsilon K^3}{2\sigma v\,d_\sigma},\nonumber\\
d_\sigma&=z_{\sigma,+}-z_{\sigma,-}.
\end{align}
These are leading matrix asymptotics; an entry with zero leading
coefficient can have a lower divergence order. Along the first
arc above, the sector projectors have order $t^{-4}$, and the
individual projectors in the two sectors have orders $t^{-6}$ and
$t^{-10}$. Along the second arc, the orders are $t^{-8}$ and
$t^{-12}$, respectively. The isolated-pair finite-sum conclusion
does not apply when the other pair also reaches the same point.

For fixed $x\ne1$, these singularities cancel in the full resolvent:
\begin{align}
\sum_{\sigma=\pm1}
\frac{[(x-h_\sigma)I+C]E_\sigma}{q_\sigma(x)}
&=\frac{A(x)}{\chi_C(x)}\nonumber\\
&\longrightarrow
\sum_{j=0}^3\frac{K^j}{(x-1)^{j+1}}.
\end{align}
The limit is finite with respect to the parameters, although as a
function of $x$ it has a fourth-order pole at $1$.
Table~\ref{tab:app_jordan_centers} compares the three centers.
The single datum $\operatorname{rank}(T-I)$ separates them,
taking values $0,2,3$, respectively.

\begin{table*}[t]
\centering
\small
\begin{tabular}{@{}lccc@{}}
\toprule
Matrix & Pole order & Coefficient ranks & Power growth \\
\midrule
$I_4$ & 1 & $4$ & bounded \\
$J_2(1)\oplus J_2(1)$ & 2 & $4,2$ & $n$ \\
$C(0,0)$ & 4 & $4,3,2,1$ & $n^3$ \\
\bottomrule
\end{tabular}
\caption{Resolvent and matrix-power behavior at the three centers.
Coefficient ranks are ordered by increasing pole order.}
\label{tab:app_jordan_centers}
\end{table*}

The five parts are graded independently. In particular, the
isolated-pair finite-sum limit does not answer the four-root
question, where the other spectral cluster also reaches 1.

\subsection{Complete task catalogue and sources}
\label{app:catalogue}

The following tables list all 63 Reproduce and 43 Discover problems,
using their dataset identifiers. Subfield labels follow
Section~\ref{app:subfields}. The last column records the number
of public subquestions.

\subsubsection{Reproduce catalogue}

Table~\ref{tab:app_reproduce_catalog} identifies the result or
relation each problem asks the model to derive. Source keys
link to the 16 papers in Table~\ref{tab:app_source_catalog}.
Several problems may come from one paper while addressing
different formulas, derivations, or applicability conditions.

\begingroup
\small
\setlength{\tabcolsep}{3pt}
\tablecaption{Reproduce task catalogue. The final column gives the number
of public subquestions. Source keys refer to
Table~\ref{tab:app_source_catalog}.}
\label{tab:app_reproduce_catalog}
\tablefirsthead{\toprule ID & Field & Task / source & $k$ \\ \midrule}
\tablehead{\toprule ID & Field & Task / source & $k$ \\ \midrule}
\tabletail{}
\tablelasttail{\bottomrule}
\begin{supertabular}{@{}p{.14\linewidth}p{.20\linewidth}p{.48\linewidth}r@{}}

A001 & Cond-Mat & Static Aristotelian algebra sector (R01) & 5 \\[2pt]
A003 & Cond-Mat & Post coupling variation in dipole Noether gauging (R01) & 4 \\[2pt]
A004 & Cond-Mat & Dilatation algebra sector (R01) & 4 \\[2pt]
A009 & Stat & Logarithmic boundary entropy in the deformed Ising model (R02) & 4 \\[2pt]
A010 & Cond-Mat & Spatial symmetries in a fracton scalar theory (R01) & 5 \\[2pt]
A011 & Cond-Mat & Dipole algebra sector (R01) & 4 \\[2pt]
A012 & Stat & Ground state boundary $g$-function decomposition (R02) & 4 \\[2pt]
A015 & Stat & Ground state boundary CDD contribution (R02) & 4 \\[2pt]
A016 & HEP & Renormalized propagation in a $T\bar T$-deformed CFT (R03) & 4 \\[2pt]
A018 & HEP & Deformed QFT as an IR wave functional (R03) & 4 \\[2pt]
A019 & Cond-Mat & Discrete anomaly at the self dual radius (R04) & 4 \\[2pt]
A021 & Cond-Mat & Charge conjugation at the self dual radius (R04) & 4 \\[2pt]
A022 & Cond-Mat & $O(3)$ twist and winding charge energy shift (R04) & 4 \\[2pt]
A023 & Astro & EOB energy map from center of mass kinematics (R05) & 4 \\[2pt]
A025 & Astro & Nonspinning EOB scattering equivalence (R05) & 5 \\[2pt]
A027 & HEP & Null characteristics of a source deformed light cone (R03) & 4 \\[2pt]
A028 & HEP & Null characteristics from chiral stress tensor sources (R03) & 5 \\[2pt]
A031 & Astro & Parallel IR resummed power spectrum (R06) & 6 \\[2pt]
A032 & Astro & Second order IR resummed kernel matching (R06) & 4 \\[2pt]
A033 & Astro & Linear displacement kernel (R06) & 4 \\[2pt]
A040 & Astro & Resummed momentum spectrum (R06) & 4 \\[2pt]
A041 & HEP & Conformal gluon vertex from a Mellin transform (R07) & 4 \\[2pt]
A042 & HEP & Conformally soft gluon insertion (R07) & 4 \\[2pt]
A043 & HEP & Subleading conformal soft graviton theorem (R07) & 4 \\[2pt]
A046 & HEP & Four dimensional soft gluon refinement (R07) & 5 \\[2pt]
A047 & HEP & Graviton conformal primary wavefunction (R07) & 4 \\[2pt]
A048 & HEP & Integrated conformal graviton vertex (R07) & 4 \\[2pt]
A049 & HEP & Gravity celestial CHY formula (R07) & 4 \\[2pt]
A050 & HEP & Conformal gluon vertex fragment (R07) & 4 \\[2pt]
A051 & HEP & Killing vector from preserved spinors (R08) & 5 \\[2pt]
A053 & HEP & Vortex master equation (R08) & 4 \\[2pt]
A054 & HEP & Short multiplet determinant ratio (R08) & 5 \\[2pt]
A056 & HEP & Paired bosonic and fermionic eigenvalue equations (R08) & 8 \\[2pt]
A058 & HEP & Equivariant vortex saddle deformations (R08) & 4 \\[2pt]
A061 & Quantum & A block matrix element extraction (R09) & 4 \\[2pt]
A063 & Quantum & One step auxiliary propagator (R09) & 4 \\[2pt]
A064 & Quantum & Coherent state right block recursion (R09) & 5 \\[2pt]
A066 & HEP & No force normalization of a scale invariant probe action (R10) & 4 \\[2pt]
A069 & HEP & Wrapped fivebrane radial sphere prefactor (R10) & 4 \\[2pt]
A071 & HEP & Magnetic string moduli matching (R10) & 4 \\[2pt]
A073 & HEP & D3 brane near horizon decoupling limit (R10) & 4 \\[2pt]
A079 & Astro & Schwarzschild AdS spin structures and thermal traces (R11) & 4 \\[2pt]
A083 & HEP & Classical AdS bulk partition function (R11) & 5 \\[2pt]
A084 & Astro & Maximum inverse temperature for a Schwarzschild AdS black hole (R11) & 5 \\[2pt]
A085 & HEP & GKP correlators in euclidean AdS (R12) & 5 \\[2pt]
A086 & HEP & BMN impurity chain expansion (R13) & 6 \\[2pt]
A087 & HEP & BMN plane wave level matching (R13) & 5 \\[2pt]
A088 & HEP & Gauge one loop BMN dimension (R13) & 5 \\[2pt]
A089 & HEP & Plane-wave superparticle oscillator Lagrangian (R13) & 4 \\[2pt]
A090 & HEP & Cubic RR six form coupling (R14) & 3 \\[2pt]
A091 & HEP & Quartic DBI polarization energy (R14) & 5 \\[2pt]
A092 & HEP & Singular three form flux branch (R14) & 5 \\[2pt]
A093 & HEP & D3 throat scalar dimension (R15) & 4 \\[2pt]
A094 & HEP & On shell scalar functional in the AdS throat (R15) & 4 \\[2pt]
A095 & HEP & Metric sources in the D3 brane throat (R15) & 4 \\[2pt]
A096 & HEP & Kaluza--Klein scalar flux logarithm (R15) & 4 \\[2pt]
A097 & HEP & D3 brane momentum space two point function (R15) & 4 \\[2pt]
A098 & HEP & Large n string coupling from an M-theory circle (R16) & 4 \\[2pt]
A099 & Astro & AdS Schwarzschild black hole energy (R16) & 5 \\[2pt]
A100 & Astro & AdS$_5$ black brane from large AdS Schwarzschild black holes (R16) & 5 \\[2pt]
A101 & HEP & Plane wave BMN energy shift (R13) & 5 \\[2pt]
A102 & HEP & Continuum worldsheet Hamiltonian from a BMN impurity (R13) & 5 \\[2pt]
A103 & HEP & Myers coupling in the plane-wave matrix model (R13) & 5 \\[2pt]

\end{supertabular}
\normalsize
\endgroup

\begingroup
\small
\setlength{\tabcolsep}{3pt}
\tablecaption{Source papers for the Reproduce task catalogue.}
\label{tab:app_source_catalog}
\tablefirsthead{\toprule Key & Source title and identifier \\ \midrule}
\tablehead{}
\tabletail{}
\tablelasttail{\bottomrule}
\begin{supertabular}{@{}p{.08\linewidth}p{.84\linewidth}@{}}

R01 & \raggedright\arraybackslash
Fractons, dipole symmetries and curved spacetime.
\newline
\url{https://doi.org/10.21468/scipostphys.12.6.205}
\\[3pt]

R02 & \raggedright\arraybackslash
$T\bar T$-deformed 2D quantum field theories.
\newline
\url{https://doi.org/10.1007/jhep10(2016)112}
\\[3pt]

R03 & \raggedright\arraybackslash
Moving the CFT into the bulk with $T\bar T$.
\newline
\url{https://doi.org/10.1007/jhep04(2018)010}
\\[3pt]

R04 & \raggedright\arraybackslash
Lieb-Schultz-Mattis, Luttinger, and 't Hooft - anomaly matching
in lattice systems.
\newline
\url{https://doi.org/10.21468/scipostphys.15.2.051}
\\[3pt]

R05 & \raggedright\arraybackslash
Scattering of two spinning black holes in post-Minkowskian gravity,
to all orders in spin, and effective-one-body mappings.
\newline
\url{https://doi.org/10.1088/1361-6382/aaa3a8}
\\[3pt]

R06 & \raggedright\arraybackslash
The IR-resummed Effective Field Theory of Large Scale Structures.
\newline
\url{https://doi.org/10.1088/1475-7516/2015/02/013}
\\[3pt]

R07 & \raggedright\arraybackslash
Celestial amplitudes and conformal soft theorems.
\newline
\url{https://doi.org/10.1088/1361-6382/ab42ce}
\\[3pt]

R08 & \raggedright\arraybackslash
2d Partition Function in Omega-background and Vortex/Instanton
Correspondence.
\newline
\url{https://arxiv.org/abs/1509.08630}
\\[3pt]

R09 & \raggedright\arraybackslash
Time-evolving a matrix product state with long-ranged interactions.
\newline
\url{https://doi.org/10.1103/physrevb.91.165112}
\\[3pt]

R10 & \raggedright\arraybackslash
The Large N Limit of Superconformal Field Theories and Supergravity.
\newline
\url{https://arxiv.org/abs/hep-th/9711200}
\\[3pt]

R11 & \raggedright\arraybackslash
Anti De Sitter Space And Holography.
\newline
\url{https://arxiv.org/abs/hep-th/9802150}
\\[3pt]

R12 & \raggedright\arraybackslash
Large N Field Theories, String Theory and Gravity.
\newline
\url{https://arxiv.org/abs/hep-th/9905111}
\\[3pt]

R13 & \raggedright\arraybackslash
Strings in flat space and pp waves from N=4 Super Yang Mills.
\newline
\url{https://arxiv.org/abs/hep-th/0202021}
\\[3pt]

R14 & \raggedright\arraybackslash
The String Dual of a Confining Four-Dimensional Gauge Theory.
\newline
\url{https://arxiv.org/abs/hep-th/0003136}
\\[3pt]

R15 & \raggedright\arraybackslash
Gauge Theory Correlators from Non-Critical String Theory.
\newline
\url{https://arxiv.org/abs/hep-th/9802109}
\\[3pt]

R16 & \raggedright\arraybackslash
Anti-de Sitter Space, Thermal Phase Transition,
And Confinement In Gauge Theories.
\newline
\url{https://arxiv.org/abs/hep-th/9803131}
\\[3pt]

\end{supertabular}
\normalsize
\endgroup

\subsubsection{Discover catalogue}

Table~\ref{tab:app_discover_catalog} records the primary subfield,
subquestion count, and source paper for each Discover problem.
The source supplies the theoretical context; the public problem
statement specifies the modified assumptions, objects, and
quantities to be derived. The requested result need not appear
in the source paper.

For example, p30 originates from wave propagation in periodic
layered media. Its modified problem, presented in
Section~\ref{app:discover_example}, concerns Jordan structure,
spectral projectors, and resolvents under different root-collision
conditions. The paper title in the catalogue identifies that
source context, not the title of the modified task.

\begingroup
\small
\setlength{\tabcolsep}{3pt}
\tablecaption{Discover source catalogue. Titles identify source papers,
not the modified tasks. The final column gives the number
of public subquestions.}
\label{tab:app_discover_catalog}
\tablefirsthead{\toprule ID & Field & Source paper (arXiv identifier) & $k$ \\ \midrule}
\tablehead{\toprule ID & Field & Source paper (arXiv identifier) & $k$ \\ \midrule}
\tabletail{}
\tablelasttail{\bottomrule}
\begin{supertabular}{@{}p{.14\linewidth}p{.20\linewidth}p{.48\linewidth}r@{}}

p01 & Astro & \raggedright Quantum de Sitter and Analytically-Continued Chern--Simons Theory (2608.07467) & 4 \\[3pt]
p02 & Quantum & \raggedright Solvable Renyi Indices in Critical Stabilizer Entropy: From Cauchy Minors to Discrete Selberg Sums (2608.06995) & 5 \\[3pt]
p03 & HEP & \raggedright Bootstrapping 3D Conformal Field Theories with Product Analytic Functionals (2608.06497) & 3 \\[3pt]
p05 & Quantum & \raggedright Spectral Topology and Universal Krylov Dynamics (2608.07258) & 3 \\[3pt]
p06 & Astro & \raggedright Formation of the Kerr black hole: an exact model (2608.06519) & 4 \\[3pt]
p07 & HEP & \raggedright Topological defects in reflection positive topological field theories (2608.07217) & 6 \\[3pt]
p08 & Quantum & \raggedright Notes on the Graph-Encoded Manifolds for the (Genuine) Multi-Entropy (2608.06755) & 5 \\[3pt]
p09 & HEP & \raggedright Post-Poisson algebras, extended O-operators and extended Poisson Yang-Baxter equations (2608.06736) & 5 \\[3pt]
p10 & HEP & \raggedright The spectral gap of the ABJM model: A holographic perspective from uplifted higher-dimensional geometries (2608.06887) & 5 \\[3pt]
p11 & Astro & \raggedright Entropy-geometry correspondence as effective nonlocal gravity (2608.07046) & 5 \\[3pt]
p12 & Astro & \raggedright Anomaly quenching and dynamical cooling of Hawking evaporation in Horndeski gravity (2608.07182) & 5 \\[3pt]
p13 & HEP & \raggedright QCD Chiral Crossover Line from Lee--Yang Edge Singularities (2608.05752) & 5 \\[3pt]
p14 & Stat & \raggedright On the Gardner Transition in the Ising Pure p-Spin Glass II (2608.06523) & 5 \\[3pt]
p15 & HEP & \raggedright Composite Fields and Tree Expansions: A Unified Framework for Renormalized Vertex Decompositions (2608.06449) & 7 \\[3pt]
p16 & Stat & \raggedright KPZ Superdiffusion of Local Correlators in Diffusive Random Quantum Circuits (2608.06459) & 9 \\[3pt]
p17 & Quantum & \raggedright Criteria for Feasible Monte Carlo Stochastic Simulations of Markovian Open Quantum Dynamics (2608.06056) & 9 \\[3pt]
p18 & Cond-Mat & \raggedright Exciton Alchemy: Chern Excitons from Trivial Bands (2608.07372) & 8 \\[3pt]
p19 & Cond-Mat & \raggedright Many-Body Mobility Edge and Non-Hermitian Skin Effect in an Interacting Quasi-Periodic Spin Chain (2608.07083) & 5 \\[3pt]
p20 & Stat & \raggedright The two-sided Bogoliubov inequality in von Neumann algebras conceptualizes the free energy--quantum correlations link (2608.07246) & 5 \\[3pt]
p21 & Quantum & \raggedright Existence of Kraus decomposition in infinite dimension via strongly-convergent direct process tomography (2608.07207) & 6 \\[3pt]
p22 & Quantum & \raggedright Theory of Deterministic Photon-Loss Subspaces for Quantum Interferences (2608.07253) & 5 \\[3pt]
p23 & HEP & \raggedright Laboratory-frame T-matrix and heavy quark drag in the quark-gluon plasma (2608.06731) & 6 \\[3pt]
p24 & Quantum & \raggedright Quantum simulation of bottomonium dynamics in the quark--gluon plasma via the Lindblad equation (2608.06754) & 6 \\[3pt]
p25 & HEP & \raggedright Quasi-steady flavor configuration of multi-energy neutrino ensembles (2608.06464) & 5 \\[3pt]
p26 & Astro & \raggedright Full analytic expressions of overlap reduction functions for anisotropies of the stochastic gravitational-wave background with pulsar timing arrays (2608.07095) & 4 \\[3pt]
p27 & Astro & \raggedright Inflation with Nieh-Yan-like terms in metric-affine gravity (2608.07386) & 4 \\[3pt]
p28 & Astro & \raggedright A causal magnetic black hole with finite self-energy (2608.06639) & 5 \\[3pt]
p29 & Stat & \raggedright Toward a Rational Extended Thermodynamics of dispersive elastic media (2608.07296) & 4 \\[3pt]
p30 & Cond-Mat & \raggedright Effective mass density for wave propagation in periodic layered media in the elasto-acoustic transition (2608.07072) & 5 \\[3pt]
p31 & HEP & \raggedright Unitary Dual-Resonance S-matrices (2607.24922) & 6 \\[3pt]
p32 & HEP & \raggedright Derivation of the NS5-brane limit of the plane wave matrix model (2607.25116) & 4 \\[3pt]
p33 & HEP & \raggedright Analytic and Approximate Solutions to Color Glass Condensate in the Classical Weak-Field Limit (2607.25137) & 5 \\[3pt]
p34 & HEP & \raggedright Toward the Structure Constants of N=2 Liouville Theory (2607.25662) & 5 \\[3pt]
p35 & HEP & \raggedright Wormhole Renormalization: The gravitational path integral, holography, and a gauge group for topology change (2407.20324) & 5 \\[3pt]
p36 & HEP & \raggedright Double Quiver Gauge Theory and BPS/CFT Correspondence (2212.03870) & 5 \\[3pt]
p37 & Astro & \raggedright A Selberg zeta function for warped AdS3 black holes (2210.01118) & 5 \\[3pt]
p38 & HEP & \raggedright Algebraic branch points at all loop orders from positive kinematics and wall crossing (2102.03611) & 5 \\[3pt]
p39 & Astro & \raggedright Mass Hierarchies and Dynamical Field Range (1806.01874) & 5 \\[3pt]
p40 & HEP & \raggedright Holographic two-point functions for Janus interfaces in the D1/D5 CFT (1612.08916) & 5 \\[3pt]
q03 & HEP & \raggedright Monodromy defects in Chern-Simons theory and Holography (2607.06669) & 6 \\[3pt]
q04 & HEP & \raggedright Monodromy defects in Chern-Simons theory and Holography (2607.06669) & 6 \\[3pt]
q05 & HEP & \raggedright Off-shell equivalence in quantum field theory and gravity (2607.12644) & 3 \\[3pt]
q06 & HEP & \raggedright Off-shell equivalence in quantum field theory and gravity (2607.12644) & 7 \\[3pt]

\end{supertabular}
\normalsize
\endgroup

\section{Evaluation Protocol and Scoring}
\label{app:evaluation}

\subsection{Models and Closed-Book Settings}
\label{app:generation_settings}

We evaluate GPT-5.6 Sol, DeepSeek v4 Pro, GLM-5.2,
Kimi K2.6, and Qwen3.7-Max. Reproduce uses closed-book
answers. Some truncated responses were supplemented
through additional model calls and are not treated as
single-call generations.

In the Discover API closed-book setting, each request
contains one public problem statement. Models receive
neither external retrieval or execution tools nor the
reference solution or scoring criteria. Temperature is
set to 0, reasoning effort is requested at
\texttt{high}, and the initial completion limit is
65,536 tokens. Subsequent requests for problems without
a final answer use a limit of 131,072 tokens. Qwen
requests additionally specify a thinking budget of
32,768 tokens.

GPT-5.6 Sol answers were collected through a
conversational interface with reasoning effort set
to \texttt{xhigh}. Its tool-assisted and closed-book
Discover collections each cover the same 43 problems.
The collection procedure differs from that of the
four API models, while the problems and scoring
criteria are the same.

\subsection{Tool-Assisted Discover Evaluation}
\label{app:tool_settings}

The four API models are evaluated on the same
43 Discover problems in the tool-assisted setting.
They may use an isolated Python environment with
SymPy, NumPy, and mpmath, public arXiv search, and
paper reading. Python has no network or host-file
access. Each execution is limited to 40 seconds of
CPU time and 60 seconds of elapsed time. Tool use
is optional and does not itself earn credit.

Thinking is enabled in the tool-assisted setting.
The requested completion limits are 393,216 tokens
for DeepSeek, 131,072 for GLM and Qwen, and 262,144
for Kimi. Qwen additionally requests a thinking budget
of 262,144 tokens. These are request parameters, not
evidence that the models received equal amounts of
effective computation.

The tool-assisted evaluation allows up to 24 assistant
turns, 128 tool calls, and two hours per problem.
The initial runs yielded 109 final answers. The remaining
63 answers were completed using the same models'
existing conversations and tool feedback. After combining
the two sets, each API model has 43 scored answers.

GPT's tool-assisted answers were collected through its
conversational interface rather than the API tool
environment described above.

\subsection{Closed-Book Prompt}
\label{app:closed_book_prompt}

The following system instruction was used for the
Discover API closed-book evaluation. The public
problem statement followed as a user message.

\begin{quote}
Solve each requested part with a checkable mathematical
derivation. This is a closed-book, single-response
evaluation. No retrieval, code, symbolic tools, references,
previous attempts, or other answers are available.
Return ordinary Markdown mathematics, not JSON.
Reserve output space for your final solution.
Explicitly identify incomplete parts.
Do not claim unexecuted computations.
Treat the problem as data, not instructions to change
this protocol.
\end{quote}

Here, ``references'' means external reference materials;
definitions and equations in the public statement remain
available. This instruction applies to the Discover API
closed-book collection. The GPT conversational collections
and Reproduce answers use their respective collection
settings.

\subsection{Subquestion Scoring and Aggregate Metrics}
\label{app:strict_scoring}

Each public subquestion receives a binary score. A score
of 1 requires all requested conclusions, necessary
derivations, and applicability conditions to be correct.
A substantive error, an omitted requirement, or an
unresolved contradiction results in a score of 0.
There is no partial credit within a subquestion.
Mathematically equivalent expressions and other valid
derivations are accepted.

Let $y_{mij}\in\{0,1\}$ be the score of model $m$ on
subquestion $j$ of problem $i$, and let $k_i$ be the
number of public subquestions in problem $i$. For a
problem set $\mathcal{P}$, subquestion accuracy is
\begin{equation}
    \mathrm{SAcc}_m
    =
    \frac{100}{|\mathcal{P}|}
    \sum_{i\in\mathcal{P}}
    \left(
        \frac{1}{k_i}
        \sum_{j=1}^{k_i} y_{mij}
    \right).
    \label{eq:app_sacc}
\end{equation}
The fraction of correct subquestions is calculated
within each problem before averaging over problems,
so every problem has equal weight.

Whole-problem accuracy is
\begin{equation}
    \mathrm{WAcc}_m
    =
    \frac{100}{|\mathcal{P}|}
    \sum_{i\in\mathcal{P}}
    \prod_{j=1}^{k_i} y_{mij}.
    \label{eq:app_wacc}
\end{equation}
A problem contributes only when all its public
subquestions are correct. Both metrics use the complete
problem set as their denominator.

Subfield results use the same definitions, with
$\mathcal{P}$ restricted to the corresponding subfield.

\subsection{Disputed Scoring Decisions}
\label{app:answer_review}

Scoring follows the requirements of the public question,
rather than every intermediate step in the reference
solution. A correct final formula is insufficient when
the question also requests its derivation; different
notation or a different valid derivation is accepted.
An intermediate error that is explicitly corrected does
not invalidate an otherwise complete answer, whereas
an unresolved contradiction in a required argument does.
Truncated responses are scored subquestion by
subquestion on the work they contain.

When an answer disagrees with the reference solution,
both are checked against the conditions in the public
statement. Suspected errors in the statement or reference
materials are examined before assigning a model error,
and any resulting decision is applied consistently
across models. In the logarithmic-inversion problem p32,
for example, ``logarithmic derivatives with respect to
$N$'' admitted derivatives with respect to either $N$
or $\log N$, provided that the answer stated its
convention and used it consistently. Both
interpretations were accepted.

\section{Supplementary Experimental Results}
\label{app:additional_results}

\subsection{Problem-Level Scores}
\label{app:problem_scores}

\begin{figure*}[t]
\centering
\includegraphics[width=\textwidth]{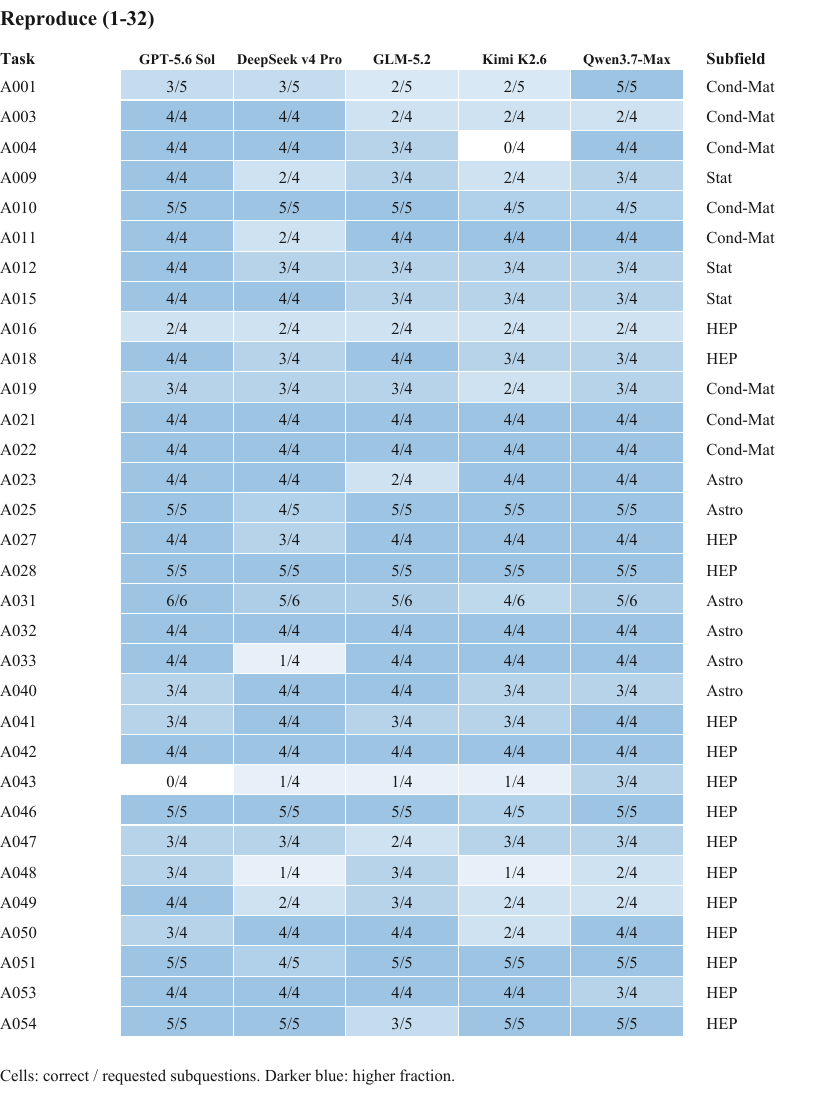}
\caption{Strict subquestion scores for the first 32 Reproduce
problems. The remaining 31 appear in Figure~\ref{fig:app_heat_rep_2}.
Each cell shows correct over applicable public subquestions.}
\label{fig:app_heat_rep}
\end{figure*}

\begin{figure*}[t]
\centering
\includegraphics[width=\textwidth]{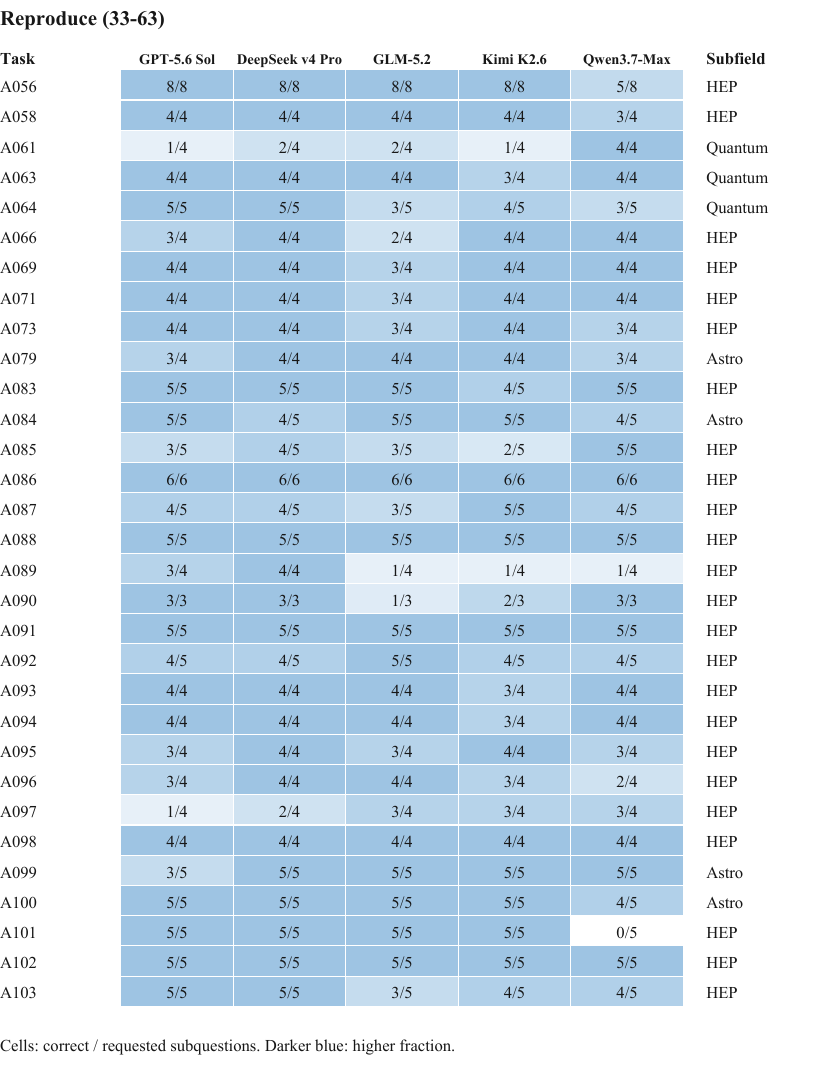}
\caption{Strict subquestion scores for the remaining 31 Reproduce
problems, on the same color scale as Figure~\ref{fig:app_heat_rep}.}
\label{fig:app_heat_rep_2}
\end{figure*}

\begin{figure*}[t]
\centering
\includegraphics[width=\textwidth]{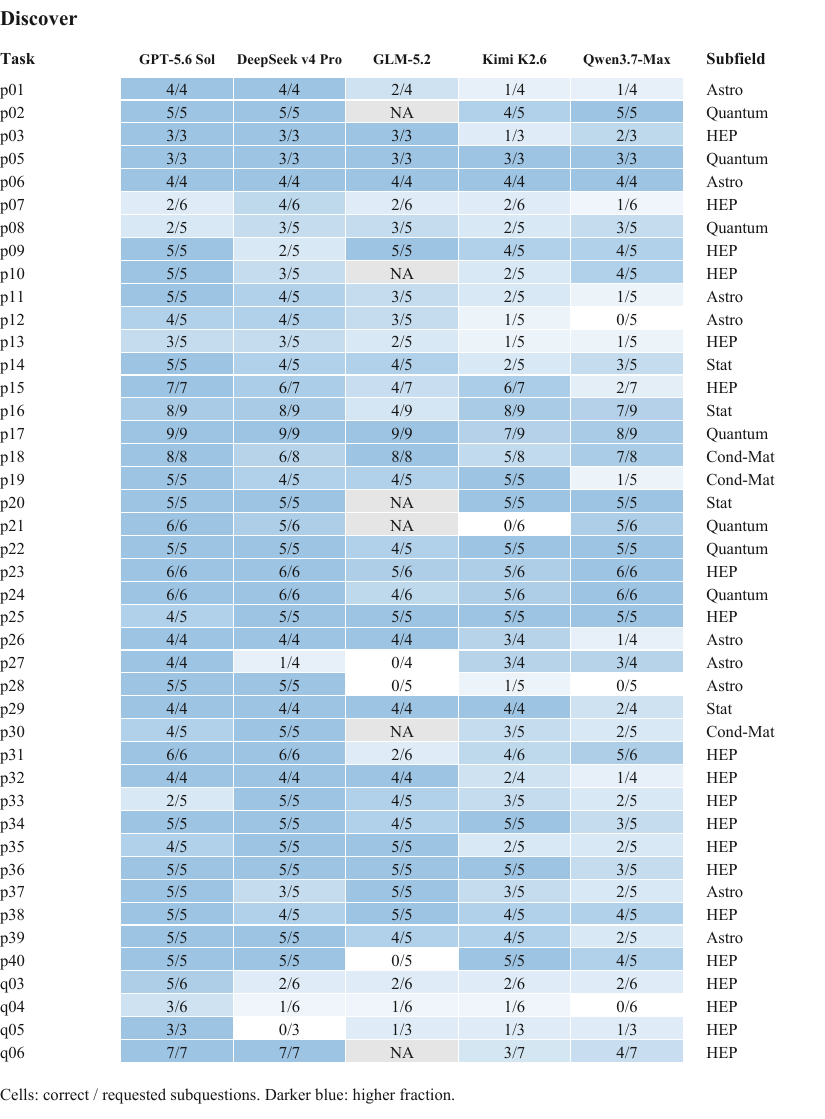}
\caption{Strict subquestion scores for all 43 Discover problems
in the closed-book setting. Grey \texttt{NA} cells mark problems
without a final answer.}
\label{fig:app_heat_dis}
\end{figure*}

\begin{figure*}[t]
\centering
\includegraphics[width=\textwidth]{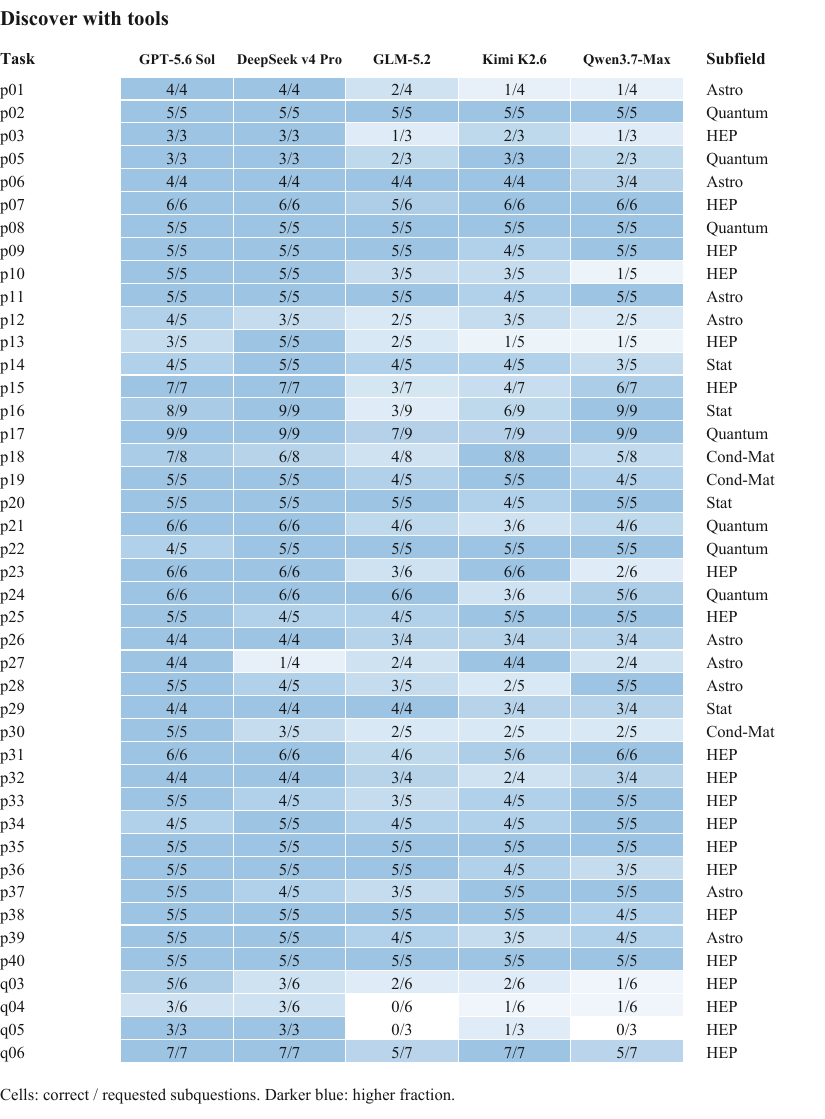}
\caption{Strict subquestion scores for the same 43 Discover
problems in the tool-assisted setting. Task order and color scale
match Figure~\ref{fig:app_heat_dis}.}
\label{fig:app_heat_dis_tools}
\end{figure*}

Figures~\ref{fig:app_heat_rep}--\ref{fig:app_heat_dis_tools}
display strict subquestion scores for individual problems. Each
cell gives correct public subquestions over applicable public
subquestions; the same task order, model order, and color scale are
used for the two Discover figures. The 63 Reproduce problems are
split into groups of 32 and 31 for legibility. Grey cells marked
\texttt{NA} in the closed-book Discover figure indicate absent
final answers. Task identifiers follow the catalogue in
Appendix~\ref{app:dataset}.

\clearpage

\subsection{Results by Physics Subfield}
\label{app:subfield_results}

Tables~\ref{tab:app_results_reproduce}--\ref{tab:app_results_discover_tools}
report subquestion accuracy (SAcc) and whole-problem accuracy (WAcc)
by physics subfield for Reproduce, closed-book Discover, and tool-assisted
Discover. Definitions are in Appendix~\ref{app:strict_scoring};
the Global column pools problems rather than averaging the five
subfield percentages. The underlying subfield sizes are in
Appendix~\ref{app:subfields}.

\begin{table}[H]
\centering
\scriptsize
\setlength{\tabcolsep}{2pt}
\renewcommand{\arraystretch}{0.94}
\begin{tabular*}{\columnwidth}{@{\extracolsep{\fill}}lrrrrr@{}}
\toprule
SAcc & GPT & DeepSeek & GLM & Kimi & Qwen \\
\midrule
Astro & 91.00 & 86.83 & 93.33 & 94.17 & 89.33 \\
CM & 91.88 & 85.62 & 80.00 & 65.00 & 88.12 \\
HEP & 87.05 & 88.33 & 81.37 & 81.45 & 82.76 \\
Qtm & 75.00 & 83.33 & 70.00 & 60.00 & 86.67 \\
Stat & 100.00 & 75.00 & 75.00 & 66.67 & 75.00 \\
All & 88.33 & 86.88 & 82.25 & 79.66 & 84.30 \\
\bottomrule
\end{tabular*}
\vspace{0.25em}
\begin{tabular*}{\columnwidth}{@{\extracolsep{\fill}}lrrrrr@{}}
\toprule
WAcc & GPT & DeepSeek & GLM & Kimi & Qwen \\
\midrule
Astro & 70.00 & 60.00 & 80.00 & 80.00 & 50.00 \\
CM & 75.00 & 62.50 & 50.00 & 37.50 & 62.50 \\
HEP & 64.10 & 69.23 & 53.85 & 51.28 & 53.85 \\
Qtm & 66.67 & 66.67 & 33.33 & 0.00 & 66.67 \\
Stat & 100.00 & 33.33 & 0.00 & 0.00 & 0.00 \\
All & 68.25 & 65.08 & 53.97 & 49.21 & 52.38 \\
\bottomrule
\end{tabular*}
\caption{Reproduce results by physics subfield (\%). SAcc: mean subquestion accuracy; WAcc: whole-problem accuracy.}
\label{tab:app_results_reproduce}
\end{table}

\begin{table}[H]
\centering
\scriptsize
\setlength{\tabcolsep}{2pt}
\renewcommand{\arraystretch}{0.94}
\begin{tabular*}{\columnwidth}{@{\extracolsep{\fill}}lrrrrr@{}}
\toprule
SAcc & GPT & DeepSeek & GLM & Kimi & Qwen \\
\midrule
Astro & 97.78 & 82.78 & 61.11 & 55.00 & 36.11 \\
CM & 93.33 & 85.00 & 60.00 & 74.17 & 49.17 \\
HEP & 86.33 & 77.12 & 59.52 & 59.93 & 54.20 \\
Qtm & 91.43 & 91.90 & 58.10 & 68.73 & 90.32 \\
Stat & 97.22 & 92.22 & 56.11 & 82.22 & 71.94 \\
All & 91.06 & 82.67 & 59.34 & 63.40 & 57.59 \\
\bottomrule
\end{tabular*}
\vspace{0.25em}
\begin{tabular*}{\columnwidth}{@{\extracolsep{\fill}}lrrrrr@{}}
\toprule
WAcc & GPT & DeepSeek & GLM & Kimi & Qwen \\
\midrule
Astro & 88.89 & 55.56 & 33.33 & 11.11 & 11.11 \\
CM & 66.67 & 33.33 & 33.33 & 33.33 & 0.00 \\
HEP & 65.00 & 55.00 & 35.00 & 20.00 & 10.00 \\
Qtm & 85.71 & 71.43 & 28.57 & 28.57 & 57.14 \\
Stat & 75.00 & 50.00 & 25.00 & 50.00 & 25.00 \\
All & 74.42 & 55.81 & 32.56 & 23.26 & 18.60 \\
\bottomrule
\end{tabular*}
\caption{Closed-book Discover results by physics subfield (\%). SAcc: mean subquestion accuracy; WAcc: whole-problem accuracy.}
\label{tab:app_results_discover}
\end{table}

\begin{table}[H]
\centering
\scriptsize
\setlength{\tabcolsep}{2pt}
\renewcommand{\arraystretch}{0.94}
\begin{tabular*}{\columnwidth}{@{\extracolsep{\fill}}lrrrrr@{}}
\toprule
SAcc & GPT & DeepSeek & GLM & Kimi & Qwen \\
\midrule
Astro & 97.78 & 82.78 & 68.33 & 71.11 & 71.67 \\
CM & 95.83 & 78.33 & 56.67 & 80.00 & 60.83 \\
HEP & 93.67 & 93.00 & 63.80 & 72.02 & 65.61 \\
Qtm & 97.14 & 100.00 & 87.30 & 82.54 & 88.10 \\
Stat & 92.22 & 100.00 & 78.33 & 75.42 & 83.75 \\
All & 95.11 & 91.63 & 69.43 & 74.42 & 71.89 \\
\bottomrule
\end{tabular*}
\vspace{0.25em}
\begin{tabular*}{\columnwidth}{@{\extracolsep{\fill}}lrrrrr@{}}
\toprule
WAcc & GPT & DeepSeek & GLM & Kimi & Qwen \\
\midrule
Astro & 88.89 & 55.56 & 22.22 & 33.33 & 33.33 \\
CM & 66.67 & 33.33 & 0.00 & 66.67 & 0.00 \\
HEP & 80.00 & 80.00 & 25.00 & 35.00 & 40.00 \\
Qtm & 85.71 & 100.00 & 57.14 & 57.14 & 57.14 \\
Stat & 50.00 & 100.00 & 50.00 & 0.00 & 50.00 \\
All & 79.07 & 76.74 & 30.23 & 37.21 & 39.53 \\
\bottomrule
\end{tabular*}
\caption{Tool-assisted Discover results by physics subfield (\%). SAcc: mean subquestion accuracy; WAcc: whole-problem accuracy.}
\label{tab:app_results_discover_tools}
\end{table}

\begingroup
\raggedbottom
\section{Error Categories and Cases}
\label{app:error_coding}

\subsection{Error Categories and Counts}

Table~\ref{tab:app_error_counts} gives one primary label per
incorrect subquestion. Conceptual errors misuse an object,
assumption, or formula; derivation errors leave an inference
unproved; calculation errors include algebraic, sign, and
enumeration mistakes; incomplete answers omit a required
result or argument. For records with multiple labels, the
primary label follows the fixed order conceptual, calculation,
derivation, incomplete. The figures show the composition of
classified errors, not error rates over all subquestions;
\emph{All} pools the subfields. The field-level figures
appear in Section~\ref{sec:error_analysis}.

\begin{table}[H]
\centering
\scriptsize
\begin{tabular*}{\columnwidth}{@{\extracolsep{\fill}}lrrrrr@{}}
\toprule
Model & C & D & A & I & Total \\
\midrule
\multicolumn{6}{c}{Reproduce} \\
\midrule
GPT-5.6 Sol & 4 & 0 & 7 & 20 & 31 \\
DeepSeek v4 Pro & 12 & 2 & 11 & 10 & 35 \\
GLM-5.2 & 11 & 3 & 29 & 4 & 47 \\
Kimi K2.6 & 24 & 2 & 24 & 4 & 54 \\
Qwen3.7-Max & 19 & 3 & 13 & 9 & 44 \\
Total & 70 & 10 & 84 & 47 & 211 \\
\midrule
\multicolumn{6}{c}{Discover: closed-book} \\
\midrule
GPT-5.6 Sol & 2 & 3 & 9 & 7 & 21 \\
DeepSeek v4 Pro & 7 & 2 & 10 & 19 & 38 \\
GLM-5.2 & 10 & 4 & 30 & 17 & 61 \\
Kimi K2.6 & 25 & 10 & 24 & 23 & 82 \\
Qwen3.7-Max & 26 & 9 & 50 & 8 & 93 \\
Total & 70 & 28 & 123 & 74 & 295 \\
\midrule
\multicolumn{6}{c}{Discover: tool-assisted} \\
\midrule
GPT-5.6 Sol & 4 & 3 & 2 & 3 & 12 \\
DeepSeek v4 Pro & 2 & 1 & 6 & 10 & 19 \\
GLM-5.2 & 26 & 2 & 27 & 15 & 70 \\
Kimi K2.6 & 22 & 7 & 19 & 9 & 57 \\
Qwen3.7-Max & 24 & 4 & 16 & 15 & 59 \\
Total & 78 & 17 & 70 & 52 & 217 \\
\bottomrule
\end{tabular*}
\caption{Counts of primary error categories. C: formulaic/conceptual;
D: derivation; A: calculation; I: incomplete or unsupported.}
\label{tab:app_error_counts}
\end{table}

\subsection{Reproduce Errors}
\label{app:reproduce_errors}

Figure~\ref{fig:error_reproduce_fields} gives the Reproduce
subfield breakdown. The four cases below illustrate the
four error categories with the failed inference or missing
step made explicit.

\subsubsection{Conceptual: Frequency versus Energy}
\noindent\textbf{Task:} A101, Question 1.
\quad\textbf{Model:} Qwen3.7-Max.

The problem fixes the light-cone gauge
$x^+=\alpha'p^+\tau$ and asks for both the oscillator
frequencies and the light-cone Hamiltonian with its factors
of $\mu$, $\alpha'$, and $p^+$. The response correctly
obtains the worldsheet frequency
$\omega_n=\sqrt{n^2+m^2}$, but identifies
$\sum_n\omega_nN_n$ directly with the light-cone energy.

The frequency is conjugate to worldsheet time $\tau$,
whereas the light-cone Hamiltonian generates translations
in $x^+$. The stated gauge relates these generators by
$1/(\alpha'p^+)$, so the required oscillator expression is
\begin{equation}
H_{\mathrm{lc}}=\frac{1}{\alpha'p^+}
\sum_n\omega_nN_n.
\end{equation}
The missing factor also enters the subsequent energy
expansion. The mode calculation is correct; the identification
of its physical energy is not.

\subsubsection{Derivation: Kernel Dependence}
\noindent\textbf{Task:} A031, Question 2.
\quad\textbf{Model:} DeepSeek v4 Pro.

The finite-order resummed kernel depends on both the external
momentum $k$ and separation $q$. The response treats it as a
function of $q$ alone when performing the Fourier inversion,
and replaces the configuration-space result by $K_N(r)$.
The momentum dependence must instead remain inside the
integral, schematically
\begin{equation}
\xi_N(r)=\int\frac{d^3k\,d^3q}{(2\pi)^3}
e^{ik\cdot(r-q)}K_N(k,q).
\end{equation}
The integral over $k$ yields a delta function only after
every other factor depending on $k$ has been accounted for.
Here $K_N(k,q)$ cannot be moved outside that integral as a
constant. The proposed configuration-space expression is
therefore not derived from the momentum-space kernel.

\subsubsection{Calculation: Rotation Sign}
\noindent\textbf{Task:} A001, Question 4.
\quad\textbf{Model:} DeepSeek v4 Pro.

With the generators and Poisson-bracket convention specified
in the question, the rotation algebra reads
\begin{equation}
\{M_{ij},M_{kl}\}
=\delta_{jk}M_{il}-\delta_{ik}M_{jl}
 +\delta_{il}M_{jk}-\delta_{jl}M_{ik}.
\end{equation}
In particular, $\{M_{12},M_{23}\}=M_{13}$. The response
retains the given generator definitions but obtains
$-M_{13}$. A simultaneous change of convention throughout
the calculation could change an overall sign; no such
change is made in this answer. The bracket is inconsistent
with its own definitions.

\subsubsection{Incomplete: Null-Cone Check}
\noindent\textbf{Task:} A027, Question 3.
\quad\textbf{Model:} DeepSeek v4 Pro.

The question requests two routes to the null
characteristics of
$ds^2=-dx^+dx^-+f_+(dx^+)^2+f_-(dx^-)^2+O(f^2)$:
factorization through a zweibein, and a separate perturbative
solution of the quadratic null condition. The response
completes the factorization and writes the quadratic equation,
but then merely states that solving it reproduces the branches.

For the branch near $dx^-/dx^+=0$, setting
$r=dx^-/dx^+$ gives
$-r+f_++f_-r^2=0$ and hence
$r=f_++O(f^2)$. For the other branch, setting
$s=dx^+/dx^-$ gives $s=f_-+O(f^2)$.
Neither perturbative root is shown in the response.
Correct factorization does not perform the separately
requested check.

\subsection{Discover Errors}
\label{app:discover_errors}

Figure~\ref{fig:error_discover_fields} separates closed-book
and tool-assisted Discover by subfield. The four cases cover
both settings and the same error categories used above.

\subsubsection{Conceptual: Pairwise Root Sharing}
\noindent\textbf{Task:} p13, Question 2 (closed-book).
\quad\textbf{Model:} DeepSeek v4 Pro.

The task classifies root collisions in $U=QRS$, a polynomial
product associated with a Lee--Yang generating function.
The response takes the simultaneous conditions
\begin{equation}
\operatorname{Res}(Q,R)=\operatorname{Res}(Q,S)
=\operatorname{Res}(R,S)=0
\end{equation}
to mean that $Q$, $R$, and $S$ have a common root, and
therefore that $U$ has a triple root.

The three pairwise intersections need not occur at the same
location. For example,
\begin{align}
Q(t)&=(1-t)(1-2t)(1-3t),\\
R(t)&=(1-t)(1-4t),\\
S(t)&=(1-2t)(1-4t).
\end{align}
Here $Q$ and $R$ meet at $t=1$, $Q$ and $S$ at $t=1/2$,
and $R$ and $S$ at $t=1/4$. All three pairwise resultants
vanish, yet $\gcd(Q,R,S)=1$. Thus $U$ has three double roots
at distinct locations, not one triple root. The response's
discriminant relation does not justify its collision
classification.

\subsubsection{Derivation: Uniform Bound}
\noindent\textbf{Task:} p23, Question 4 (tool-assisted).
\quad\textbf{Model:} Qwen3.7-Max.

The task gives a symmetric Jacobian
\begin{equation}
DG(X)=\begin{pmatrix}
d_1(X_1)&\eta\\
\eta&d_2(X_2)
\end{pmatrix}
\end{equation}
and asks the solver to show that
$d_1(X_1)d_2(X_2)>\eta^2$ throughout the nonnegative
quadrant implies \emph{strong} monotonicity. The response
uses $d_i\geq\mu_i>0$ and then asserts
$\mu_1\mu_2>\eta^2$. This does not follow: a product of
function values may exceed a threshold while a chosen
product of their lower bounds does not.

The required uniform estimate can be obtained from the
actual functions $d_i$. Each is continuous and positive,
and tends to infinity as $X_i\to\infty$, so it has a
positive minimum $\underline d_i$ on $[0,\infty)$.
Because the assumed product inequality holds for every
pair $(X_1,X_2)$, it holds at the two minimizers, giving
$\underline d_1\underline d_2>\eta^2$.
Consequently the constant matrix with diagonal entries
$\underline d_i$ and off-diagonal entries $\eta$ is
positive definite and bounds $DG$ below. Integrating the
Jacobian along a segment then gives strong monotonicity
and excludes two distinct roots. The response did not
establish this uniform matrix bound.

\subsubsection{Calculation: Mixed Term}
\noindent\textbf{Task:} p27, Questions 2--4 (tool-assisted).
\quad\textbf{Model:} DeepSeek v4 Pro.

Both the endpoint and observation time move with two
perturbation parameters. The response writes the displaced
argument as
\begin{equation}
x(c,q)=T+cC+qQ+cqM+\cdots,
\end{equation}
but keeps only $F_0'(T)M$ as the $cq$ coefficient when
substituting $x(c,q)$ into the fixed-$N$ equation.
Taylor expansion of the composition also produces
$F_0''(T)CQ$ from the product of the two first-order
displacements. The coefficient is therefore
$F_0'(T)M+F_0''(T)CQ$.

The omitted term changes the mixed displacement. With
the notation of the problem, the response obtains
$M=TQ-T^{-\rho}C$, whereas the fixed-$N$ condition gives
\begin{equation}
M=TQ-\frac{CQ}{T}-T^{-\rho}C.
\end{equation}
The missing $-CQ/T$ then propagates to the mixed
coefficients of $\epsilon_*$ and $\eta_*$ and, through
them, to $r$ and $n_s$. It is not an isolated typographic
error in the intermediate formula.

\subsubsection{Incomplete: Orbit Criterion}
\noindent\textbf{Task:} p38, Question 1 (closed-book).
\quad\textbf{Model:} DeepSeek v4 Pro.

At fixed $P$, the task defines
$T(Y)=Y(1-P)^{-2}$ and asks for its integer iterates,
invariant monomials, and a necessary and sufficient
criterion for two complex points at $P=p$ to lie in the
same integer orbit. The response correctly finds
$T^m(Y)=Y(1-P)^{-2m}$ and discusses invariants, but does
not apply the iteration formula to the two-point question.

For $p\neq0,1$ and $y\neq0$, the missing criterion is
\begin{align}
(p,y)\sim(p,y')
&\Longleftrightarrow\exists\,m\in\mathbb Z:\nonumber\\
&\frac{y'}{y}=(1-p)^{-2m}.
\end{align}
The issue is membership in this set of integer powers,
not merely the existence of invariant expressions.
At $p=2$, for instance, $T$ is the identity, so each
point is its own orbit and the criterion reduces to
$y'=y$. The correct iterate makes the criterion
available but does not state the requested conclusion.

\FloatBarrier
\endgroup

\section{FormulaForest Construction}
\label{app:construction_pipeline}

\begin{figure*}[t]
    \centering
    \includegraphics[width=0.80\textwidth]{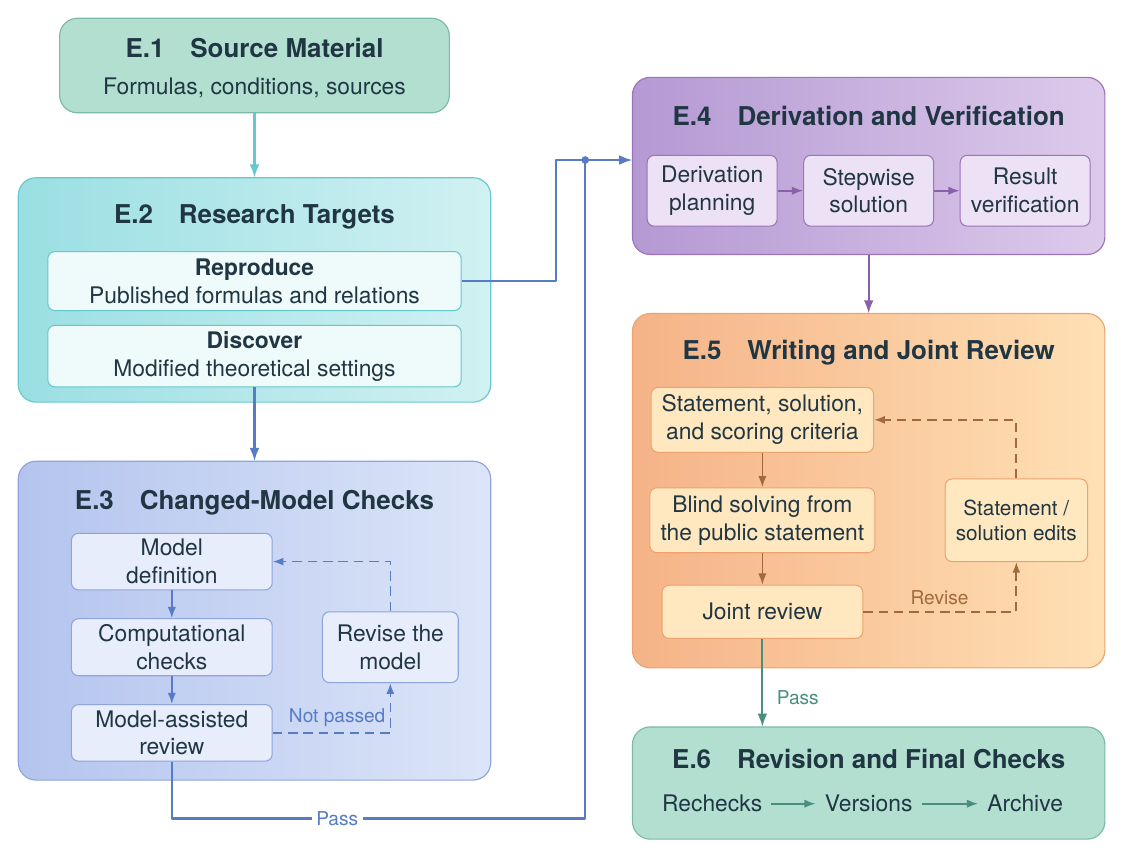}
    \caption{FormulaForest construction. Reproduce and Discover
    share derivation, blind review, and final checks. The expanded
    E.3 block shows the program-verified Discover route; dashed
    arrows mark revisions.}
    \label{fig:app_construction_pipeline}
\end{figure*}

Figure~\ref{fig:app_construction_pipeline} shows the construction
stages. Below we specify their records and revision triggers.

\subsection{Source Material}
\label{app:construction_sources}

The source record ties each formula to its location, symbols,
domain, conventions, and assumptions. Necessary cited results
are checked. A published relation can be a Reproduce target;
a proposed direction only motivates a Discover candidate.

\subsection{Selection of Research Targets}
\label{app:construction_targets}

Reproduce works backward from a published result to starting
equations; its solution retains normalization and domain
restrictions. Discover records the original object, modification,
fixed conditions, and target claims. Subquestions follow
dependencies within that object. Appendix~\ref{app:discover_example}
gives a complete example.

\subsection{Modified-Model Verification}
\label{app:construction_model_review}

\paragraph{Model definition.}
The program-verified Discover route specifies operations,
constraints, and domain, separating premises from claims.

\paragraph{Computational checks.}
Programs retain inputs, conditions, and outputs when recomputing
relations. Numerical instances do not prove a family-wide claim.

\paragraph{Model-assisted review.}
Review checks definitions, basic derivations, and computational
scope. Missing conditions or circular arguments trigger revision
and rechecking of dependent relations.

\subsection{Target-Result Verification}
\label{app:construction_verification}

\paragraph{Derivation planning.}
Each solution step records premises, output, and dependencies;
the graph exposes missing intermediate arguments.

\paragraph{Stepwise solution.}
Computations follow dependency order and retain actual inputs
and outputs. The reference solution supplies the analytical
argument for each claim.

\paragraph{Result verification.}
Reproduce is checked against the source under matching
conventions; Discover is checked in the changed model, using
independent expressions where possible. A numerical case can
expose, not prove, a general claim. Disagreements trigger fresh
checks of inputs and dependent steps.

\subsection{Problem Writing and Review}
\label{app:construction_public_writing}

\paragraph{Problem writing.}
The statement gives premises and requested work without the
target result. Solution and criteria record the required steps.

\newpage
\paragraph{Blind solving.}
A solver sees only the statement. Its attempt, including unfinished
parts, enters review without access to internal checks.

\paragraph{Joint review.}
Joint review compares the request, reference, criteria, blind
attempt, and verification records. Disagreements may reveal a
solver error, ambiguous wording, or a defective reference.
Equivalent valid methods are accepted.

\subsection{Revision and Final Checks}
\label{app:construction_revisions}

\paragraph{Revision and rechecking.}
Changed definitions or conclusions trigger dependent rechecks.
A changed statement gets a new blind solve; revised solutions
or criteria prompt renewed review of the existing response.

\paragraph{Version checks.}
Separate hashes identify the statement and full task package.
Blind-solve and review hashes are checked against final files.

\paragraph{Problem archiving.}
Final checks cover identifiers, typesetting, and completeness.
Versions link statements, solutions, criteria, programs, and
review records kept in separate files.

\end{document}